\documentclass{llncs}
\usepackage{llncsdoc}
\usepackage{makeidx}

\usepackage{url,mathrsfs}
\usepackage{amsmath}
\usepackage{amsfonts}
\usepackage{subfigure}
\usepackage{graphicx}

\title{\large Computationally Efficient Estimators for Dimension Reductions Using Stable Random Projections\vspace{-0.2in}}
\titlerunning{Efficient Stable Random Projections}
\author{Ping Li\vspace{-0.1in}}
\authorrunning{Ping Li}   
\tocauthor{Ping Li (Cornell University),}
\institute{Department of Statistical Science, Cornell University, Ithaca NY 14853, USA\\
\email{\{pingli\}@cornell.edu}\vspace{-0.3in}}

\begin{document}

\maketitle

\begin{abstract}
The\footnote{First draft Feb. 2008, slightly revised in June 2008. The results were announced in January 2008 at SODA'08 when the author presented the work of \cite{Proc:Li_SODA08}.} method of {\em stable random projections} is a  tool for efficiently computing the $l_\alpha$   distances using low memory, where $0<\alpha \leq 2$ is a tuning parameter. The method boils down to a statistical estimation task and various estimators have been proposed, based on the {\em geometric mean}, the {\em harmonic mean}, and the {\em fractional power} etc.

This study proposes the \textbf{\em optimal quantile} estimator, whose main operation is \textbf{\em selecting}, which is considerably less expensive than taking fractional power, the main operation in previous estimators. Our experiments report that the {\em optimal quantile} estimator is nearly one order of magnitude more computationally efficient than previous estimators. For  large-scale learning  tasks in which storing and computing pairwise distances is a serious bottleneck, this estimator should be desirable.

In addition to its computational advantages, the {\em optimal quantile} estimator exhibits nice theoretical properties. It is more accurate than previous estimators when {\small$\alpha>1$}. We  derive its theoretical error bounds and  establish the explicit (i.e., no hidden constants) sample complexity bound.

\end{abstract}

\vspace{-0.4in}
\section{Introduction} \label{sec_intro}
\vspace{-0.1in}

The method of {\em stable random projections}\cite{Article:Indyk_JACM06,Proc:Li_SODA08,Proc:Li_Hastie_NIPS07}, as an efficient tool for computing pairwise distances in massive high-dimensional data,  provides a promising mechanism to tackle  some of the challenges in modern machine learning. In this paper, we provide an easy-to-implement algorithm for {\em stable random projections} which is both statistically accurate and computationally efficient.

\vspace{-0.15in}
\subsection{Massive High-dimensional Data in Modern Machine Learning}
\vspace{-0.05in}

We denote a data matrix  by $\mathbf{A}\in\mathbb{R}^{n\times D}$, i.e., $n$ data points in $D$ dimensions. Data sets in modern applications exhibit important characteristics which impose tremendous challenges in machine learning \cite{Book:Bottou_07}:
\begin{itemize}
\item Modern data sets with $n=10^5$ or even $n=10^6$   points  are not uncommon in supervised learning, e.g., in image/text classification, ranking algorithms for search engines, etc. In the unsupervised domain (e.g., Web clustering, ads clickthroughs, word/term associations), $n$ can be even much larger.
\item Modern data sets are often of ultra high-dimensions ($D$), sometimes in the order of millions (or even higher), e.g., image, text, genome (e.g., SNP), etc. For example, in image analysis, $D$ may be $10^3\times 10^3 = 10^6$ if using pixels as features, or $D = 256^3 \approx 16$ million if using color histograms as features.
\item Modern data sets are sometimes collected in a dynamic streaming fashion.
\item Large-scale data  are often heavy-tailed, e.g., image and text data.
\end{itemize}

Some large-scale data  are  dense, such as image and genome data.  Even for  data sets which are  sparse, such as text, the absolute number of non-zeros may be still large. For example, if one queries ``machine learning" (a not-too-common term) in Google.com, the total number of pagehits is about 3 million. 
In other words, if one builds a term-doc matrix at Web scale, although the matrix is  sparse, most rows will  contain large numbers (e.g., millions) of non-zero entries.

\vspace{-0.1in}
\subsection{Pairwise Distances in Machine Learning}
\vspace{-0.05in}

Many learning algorithms require a similarity matrix computed from pairwise distances of the data matrix $\mathbf{A}\in\mathbb{R}^{n\times D}$. Examples include clustering, nearest neighbors, multidimensional scaling, and kernel SVM (support vector machines). The similarity matrix requires $O(n^2)$ storage space and $O(n^2D)$ computing time.

This study focuses on the $l_\alpha$ distance ($0<\alpha\leq 2$). Consider two vectors $u_1$, $u_2\in\mathbb{R}^D$ (e.g., the leading two rows in $\mathbf{A}$), the $l_\alpha$ distance between $u_1$ and $u_2$ is\vspace{-0.15in}

{\small\begin{align}
d_{(\alpha)} = \sum_{i=1}^D|u_{1,i}-u_{2,i}|^\alpha.
\end{align}}\vspace{-0.15in}

Note that, strictly speaking, the $l_\alpha$ distance should be defined as {\small$d^{1/\alpha}_{(\alpha)}$}. Because the power operation {\small$(.)^{1/\alpha}$} is the same for all pairs, it often makes no difference whether we use {\small$d^{1/\alpha}_{(\alpha)}$} or just $d_{(\alpha)}$; and hence we focus on $d_{(\alpha)}$.

The  radial basis kernel (e.g., for SVM) is constructed from $d_{(\alpha)}$ \cite{Article:Chapelle_99,Book:Scholkopf_02}:\vspace{-0.15in}

{\small\begin{align}\label{eqn_rbs_kernel}
\mathbf{K}(u_1,u_2) = \exp\left(-\gamma\sum_{i=1}^D\left|u_{1,i} - u_{1,i}\right|^\alpha\right), \hspace{0.2in} 0<\alpha\leq 2.
\end{align}}\vspace{-0.15in}

When $\alpha=2$, this is the Gaussian radial basis kernel.  Here $\alpha$ can be viewed as a {\em tuning} parameter. For example, in their histogram-based image classification project using SVM, \cite{Article:Chapelle_99} reported that $\alpha = 0$ and $\alpha = 0.5$ achieved  good performance.
For heavy-tailed data, tuning $\alpha$  has the similar effect as term-weighting the original data, often a critical step in a lot of applications \cite{Article:Leopold_ML02,Proc:Rennie_ICML03}.

For  popular kernel SVM solvers including the  {\em Sequential Minimal Optimization (SMO}) algorithm\cite{Proc:Platt_NIPS98}, storing and computing kernels is the major bottleneck. Three  computational challenges were summarized in \cite[page 12]{Book:Bottou_07}:
\begin{itemize}
\item \textbf{\em Computing kernels is expensive}
\item \textbf{\em Computing full kernel matrix is wasteful} \hspace{0.3in}
 Efficient SVM solvers often do not need to evaluate all pairwise kernels.
\item \textbf{\em Kernel matrix does not fit in memory}\hspace{0.3in}
Storing the kernel matrix  at the memory cost {\small$O(n^2)$} is challenging when {\small $n>10^5$}, and is not realistic for {\small$n>10^6$}, because {\small$O\left(10^{12}\right)$} consumes at least $1000$ GBs memory.
\end{itemize}

A popular strategy in large-scale learning is to evaluate distances \textbf{on the fly}\cite{Book:Bottou_07}. That is, instead of loading the similarity matrix in  memory at the cost of $O(n^2)$, one can load the original data matrix at the cost of $O(nD)$ and recompute  pairwise distances on-demand. This strategy is apparently problematic when $D$ is not too small. For high-dimensional data, either loading the data matrix in memory is unrealistic or computing distances on-demand becomes too expensive.

Those challenges are not unique to kernel SVM; they are general issues in distanced-based learning algorithms. The method of {\em stable random projections} provides a promising scheme by reducing the dimension $D$ to a small $k$ (e.g., $k = 50$), to facilitate compact data storage and efficient distance computations.

\vspace{-0.1in}
\subsection{Stable Random Projections}
\vspace{-0.05in}

The basic procedure of {\em stable random projections} is to multiply $\mathbf{A}\in\mathbb{R}^{n\times D}$ by a random  matrix $\mathbf{R}\in\mathbb{R}^{D\times k}$ ($k\ll D$), which is generated by sampling each entry $r_{ij}$ i.i.d. from a  symmetric stable distribution $S(\alpha,1)$. The resultant  matrix $\mathbf{B} = \mathbf{A\times R}\in\mathbb{R}^{n\times k}$ is much smaller than $\mathbf{A}$ and hence it may fit in memory.

Suppose a stable random variable $x \sim S(\alpha, d)$, where $d$ is the scale parameter.  Then its
characteristic function (Fourier transform of the density function) is\vspace{-0.15in}

{\small\begin{align}\notag
\text{E}\left(\exp\left(\sqrt{-1}x t\right)\right) =
\exp\left(-d|t|^\alpha\right),
\end{align}}\vspace{-0.15in}

\noindent which does not have a closed-form inverse except for
$\alpha = 2$ (normal) or $\alpha = 1$ (Cauchy). Note that when $\alpha = 2$, $d$ corresponds to  ``$\sigma^2$''  (not ``$\sigma$'') in a normal.

Corresponding to the leading two rows in  $\mathbf{A}$, $u_1$, $u_2\in\mathbb{R}^D$, the leading two rows in $\mathbf{B}$ are $v_1 = \mathbf{R}^\text{T}u_1$, $v_2 = \mathbf{R}^\text{T}u_2$. The entries of the difference, \vspace{-0.15in}

{\small\begin{align}\notag
x_j = v_{1,j} - v_{2,j} = \sum_{i=1}^D r_{ij}\left(u_{1,i} -u_{2,i}\right) \sim S\left(\alpha, d_{(\alpha)} = \sum_{i=1}^D|u_{1,i} - u_{2,i}|^\alpha\right),
\end{align}}\vspace{-0.1in}

\noindent for $j = 1$ to $k$, are i.i.d. samples from a stable distribution with the scale parameter being the $l_\alpha$ distance $d_{(\alpha)}$, due to properties of Fourier transforms. For example, when $\alpha = 2$, a weighted sum of i.i.d. standard normals is also normal with the scale parameter (i.e., variance) being the sum of squares of all weights.

Once we obtain the stable samples,  one can discard the original  matrix $\mathbf{A}$ and the remaining task is to estimate the scale parameter $d_{(\alpha)}$ for each pair.\\ 

Some applications of {\em stable random projections} are summarized as follows:\vspace{-0.05in}
\begin{itemize}
\item \textbf{\em Computing all pairwise distances} \hspace{0.3in} The cost of computing all pairwise distances of
  {\small$\mathbf{A}\in\mathbb{R}^{n\times D}$, $O(n^2D)$}, is significantly
  reduced to {\small$O(nDk + n^2k)$}.
\item \textbf{\em Estimating $l_\alpha$ distances online} \hspace{0.3in}
  For {\small$n>10^5$}, it is challenging or unrealistic to materialize all pairwise
  distances in $\mathbf{A}$. Thus, in applications such as online learning, databases,
  search engines, and online recommendation systems, it is often more efficient if we
store $\mathbf{B}\in\mathbb{R}^{n\times k}$ in the memory and estimate any
 distance  {\em on the fly}  if needed. Estimating distances online is the standard strategy in large-scale kernel learning\cite{Book:Bottou_07}. With {\em stable random projections}, this simple strategy becomes effective in high-dimensional data.
\item \textbf{\em Learning with dynamic streaming data} \hspace{0.3in} In reality, the data matrix may be updated overtime.  In fact, with streaming data arriving at high-rate\cite{Article:Indyk_JACM06,Proc:Babcock_PODS02}, the ``data matrix'' may be never stored and hence all operations (such as clustering and classification) must be conducted on the fly. The method of {\em stable random projections} provides a scheme to compute and update  distances on the fly in one-pass of the data; see relevant papers  (e.g., \cite{Article:Indyk_JACM06}) for more details on this important and fast-developing subject.
\item \textbf{\em Estimating entropy} \hspace{0.08in} The entropy distance {\small$\sum_{i=1}^D|u_{1,i} - u_{2,i}|\log |u_{1,i} - u_{2,i}|$} is a  useful  statistic. A workshop in NIPS'03  {\small(\url{www.menem.com/~ilya/pages/NIPS03})} focused on entropy estimation. A recent practical algorithm is simply using the difference between the $l_{\alpha_1}$ and  $l_{\alpha_2}$ distances\cite{Proc:Zhao_IMC07}, where {\small$\alpha_1 = 1.05$, $\alpha_2=0.95$}, and the distances were estimated by {\em stable random projections}.
\end{itemize}

If one tunes the $l_\alpha$ distances for many different $\alpha$ (e.g.,  \cite{Article:Chapelle_99}), then  {\em stable random projections} will be even more desirable as a cost-saving device.

\vspace{-0.1in}
\section{The Statistical Estimation Problem}
\vspace{-0.05in}

Recall that the method of {\em stable random projections} boils down to a statistical estimation problem. That is, estimating the scale parameter $d_{(\alpha)}$ from $k$ i.i.d. samples $x_j \sim S(\alpha, d_{(\alpha)})$, $j = 1$ to $k$. We consider that a good estimator $\hat{d}_{(\alpha)}$ should have the following desirable properties:
\begin{itemize}
\item (Asymptotically) unbiased and small variance.
\item Computationally efficient.
\item Exponential decrease of error (tail) probabilities.
\end{itemize}

The {\em arithmetic mean} estimator {\small$\frac{1}{k}\sum_{j=1}^k|x_j|^2$} is  good for $\alpha =2$. When $\alpha<2$, the task is less straightforward  because (1) no explicit density of $x_j$ exists unless $\alpha = 1$ or $0+$; and (2) $\text{E}(|x_j|^t)<\infty$ only when $-1<t<\alpha$.

\vspace{-0.1in}
\subsection{Several Previous Estimators}

Initially reported in arXiv in 2006,  \cite{Proc:Li_SODA08} proposed the {\em geometric mean} estimator \vspace{-0.05in} {\small\begin{align}\notag
\hat{d}_{(\alpha),gm} &= \frac{\prod_{j=1}^k
  |x_j|^{\alpha/k}}{\left[\frac{2}{\pi}\Gamma\left(\frac{\alpha}{k}\right)\Gamma\left(1-\frac{1}{k}\right)\sin\left(\frac{\pi}{2}\frac{\alpha}{k}\right)\right]^k}.
\end{align}}
where $\Gamma(.)$ is the Gamma function, and the {\em harmonic mean} estimator\vspace{-0.06in}
{\small\begin{align}\notag
\hat{d}_{(\alpha),hm}
&=\frac{-\frac{2}{\pi}\Gamma(-\alpha)\sin\left(\frac{\pi}{2}\alpha\right)}{\sum_{j=1}^k|x_j|^{-\alpha}}\left(k - \left(\frac{-\pi\Gamma(-2\alpha)\sin\left(\pi\alpha\right)}{\left[\Gamma(-\alpha)\sin\left(\frac{\pi}{2}\alpha\right)\right]^2}-1
\right)\right).
\end{align}}
More recently, \cite{Proc:Li_Hastie_NIPS07} proposed the {\em fractional power} estimator\vspace{-0.1in}
{\small\begin{align}\notag
&\hat{d}_{(\alpha),fp}
= \left(\frac{1}{k}\frac{\sum_{j=1}^k|x_j|^{\lambda^*
    \alpha}}{\frac{2}{\pi}
  \Gamma(1-\lambda^*)\Gamma(\lambda^*\alpha)\sin\left(\frac{\pi}{2}\lambda^*\alpha\right)} \right)^{1/\lambda^*} \times \\\notag
&\hspace{0.in} \left(1-\frac{1}{k}\frac{1}{2\lambda^*}\left(\frac{1}{\lambda^*}-1\right)\left(\frac{\frac{2}{\pi}
  \Gamma(1-2\lambda^*)\Gamma(2\lambda^*\alpha)\sin\left(\pi\lambda^*\alpha\right)}{\left[\frac{2}{\pi}
  \Gamma(1-\lambda^*)\Gamma(\lambda^*\alpha)\sin\left(\frac{\pi}{2}\lambda^*\alpha\right)\right]^2}-1  \right) \right),
\end{align}}
\noindent where\vspace{-0.2in}
{\small\begin{align}\notag
\lambda^* =
\underset{-\frac{1}{2\alpha}\lambda<\frac{1}{2}}{\text{argmin}}\ \
 \frac{1}{\lambda^2}\left(\frac{\frac{2}{\pi}
  \Gamma(1-2\lambda)\Gamma(2\lambda\alpha)\sin\left(\pi\lambda\alpha\right)}{\left[\frac{2}{\pi}
  \Gamma(1-\lambda)\Gamma(\lambda\alpha)\sin\left(\frac{\pi}{2}\lambda\alpha\right)\right]^2}-1  \right).
\end{align}}\vspace{-0.1in}

\begin{figure}[h]\vspace{-0.12in}
\begin{center}
\includegraphics[width = 2.5in]{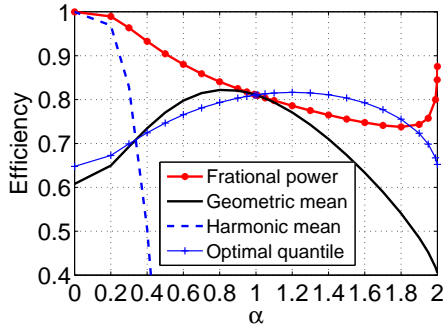}
\end{center}
\vspace{-0.3in}
\caption{The Cram\'er-Rao efficiencies (the higher the better, max = $100\%$) of various estimators, including the {\em optimal quantile} estimator proposed in this study.
}\label{fig_efficiency}\vspace{-0.15in}
\end{figure}

All three estimators are unbiased or asymptotically (as {\small$k\rightarrow\infty$}) unbiased. Figure \ref{fig_efficiency} compares their asymptotic variances in terms of the Cram\'er-Rao efficiency, which is the ratio of the smallest possible asymptotic   variance over the asymptotic variance of the estimator, as {\small$k\rightarrow\infty$}.

The {\em geometric mean} estimator, $\hat{d}_{(\alpha),gm}$ exhibits tail bounds in exponential forms, i.e., the errors decrease exponentially fast: \vspace{-0.15in}

{\small\begin{align}\notag
\mathbf{Pr}\left(|\hat{d}_{(\alpha),gm} - d_{(\alpha)}| \geq \epsilon d_{(\alpha)}\right) \leq 2\exp\left(-k\frac{\epsilon^2}{G_{gm}}\right).
\end{align}}\vspace{-0.15in}

The {\em harmonic mean} estimator, $\hat{d}_{(\alpha),hm}$, works well for small $\alpha$, and has exponential tail bounds for  $\alpha=0+$.

The {\em fractional power} estimator, $\hat{d}_{(\alpha),fp}$,  has smaller asymptotic variance than both the {\em geometric mean} and  {\em harmonic mean} estimators. However, it does not have exponential tail bounds, due to the restriction  {\small$-1<\lambda^*\alpha<\alpha$} in its definition.  As shown in \cite{Proc:Li_Hastie_NIPS07}, it only has finite moments slightly higher than the $2nd$ order, when $\alpha$ approaches 2 (because {\small$\lambda^*\rightarrow 0.5$}), meaning that large errors may have a good chance to occur. We will demonstrate this by simulations.

\vspace{-0.1in}
\subsection{The Issue of Computational Efficiency}
\vspace{-0.05in}

In the definitions of $\hat{d}_{(\alpha),gm}$, $\hat{d}_{(\alpha),hm}$ and $\hat{d}_{(\alpha),fp}$,  all three estimators require evaluating fractional powers, e.g., $|x_j|^{\alpha/k}$. This operation is relatively expensive, especially if we need to conduct this tens of billions of times (e.g., $n^2 =10^{10}$).

For example, \cite{Article:Chapelle_99} reported that, although the radial basis kernel (\ref{eqn_rbs_kernel}) with $\alpha = 0.5$ achieved good performance, it was not preferred because evaluating the square root was too expensive.

\vspace{-0.1in}
\subsection{Our Proposed Estimator}
\vspace{-0.05in}

We propose the {\em optimal quantile} estimator, using the $q^*$th smallest {\small$|x_j|$:
\begin{align}\label{eqn_d_oq}
\hat{d}_{(\alpha),oq}\propto \left(q^*\text{-quantile}\{|x_j|,j=1, 2, ..., k\}\right)^\alpha,
 \end{align}
where $q^*=q^*(\alpha)$} is chosen to minimize the asymptotic variance.

This estimator is computationally attractive because \textbf{selecting} should be much less expensive than evaluating fractional powers. If we are interested in {\small$d_{(\alpha)}^{1/\alpha}$} instead, then we do not even need to evaluate any fractional powers.

As  mentioned, in many cases using either $d_{(\alpha)}$ or  {\small$d_{(\alpha)}^{1/\alpha}$} makes no difference and $d_{(\alpha)}$ is often preferred because it avoids taking {\small$(.)^{1/\alpha}$} power. The radial basis kernel (\ref{eqn_rbs_kernel}) requires $d_{(\alpha)}$. Thus this study focuses on $d_{(\alpha)}$. On the other hand, if we can estimate {\small$d_{(\alpha)}^{1/\alpha}$} directly, for example, using (\ref{eqn_d_oq}) without  the $\alpha$th power, we might as well just use {\small$d_{(\alpha)}^{1/\alpha}$}  if permitted. In case we do not  need to  evaluate any fractional power, our estimator will be even more computationally efficient.

In addition to the computational advantages, this estimator also has good theoretical properties,  in terms of  both the variances and tail probabilities:
\begin{enumerate}
\item Figure \ref{fig_efficiency} illustrates that, compared with the {\em geometric mean} estimator, its asymptotic variance is about the same when $\alpha <1$, and is considerably smaller when $\alpha>1$. Compared with the {\em fractional power} estimator, it has smaller asymptotic variance when $1<\alpha \leq 1.8$. In fact, as will be shown by simulations, when the sample size $k$ is not too large, its mean square errors are considerably smaller  than the {\em fractional power} estimator when $\alpha >1$.
\item The {\em optimal quantile} estimator exhibits tail bounds in exponential forms. This theoretical contribution is practically important, for selecting the sample size $k$. In  learning theory, the generalization bounds are often loose. In our case, however, the bounds are tight because the distribution is  specified.
\end{enumerate}

The next section will be devoted to analyzing the {\em optimal  quantile} estimator.

\vspace{-0.15in}
\section{The Optimal Quantile Estimator}
\vspace{-0.05in}

Recall  the goal is to estimate
$d_{(\alpha)}$ from
{\small$\{x_j\}_{j=1}^k$}, where {\small$x_j\sim S(\alpha,d_{(\alpha)})$}, i.i.d. Since the distribution belongs to the scale family, one can estimate the scale parameter from quantiles. Due to
 symmetry, it is natural to consider the absolute values:\vspace{-0.1in}

{\small\begin{align}\label{eqn_quantile}
\hat{d}_{(\alpha),q} = \left(\frac{q\text{-Quantile}\{|x_j|, j = 1, 2, ..., k\}}{q\text{-Quantile}\{|S(\alpha,1)|\}}\right)^\alpha,
\end{align}\vspace{-0.1in}
}

\noindent which is best understood by the fact that if {\small$x \sim S(\alpha, 1)$}, then {\small$d^{1/\alpha}x \sim S(\alpha, d)$}, or more obviously, if $x\sim N(0,1)$, then {\small$\left(\sigma^2\right)^{1/2} x\sim N\left(0,\sigma^2\right)$}. By properties of order statistics \cite{Book:David},
any $q$-quantile will provide an asymptotically unbiased estimator.

Lemma \ref{lem_var_q} provides the asymptotic variance of $\hat{d}_{(\alpha),q}$.
\begin{lemma}\label{lem_var_q}
Denote {\small$f_X\left(x; \alpha,d_{(\alpha)}\right)$} and {\small$F_X\left(x; \alpha,d_{(\alpha)}\right)$} the probability density function and the cumulative density function  of {\small$X\sim S(\alpha, d_{(\alpha)})$}, respectively.

The asymptotic variance of $\hat{d}_{(\alpha),q}$ defined in (\ref{eqn_quantile}) is
{\small\begin{align}\label{eqn_var_q}
\text{Var}\left(\hat{d}_{(\alpha),q}\right)
=&\frac{1}{k}\frac{(q-q^2)\alpha^2/4}{f^2_X\left(W;\alpha,1\right) W^2 } d_{(\alpha)}^2+O\left(\frac{1}{k^2}\right)
\end{align}}
\noindent where {\small$W = F_X^{-1}\left((q+1)/2;\alpha,1\right)=q\text{-Quantile}\{|S(\alpha,1)|\}$}.

\textbf{Proof:} See Appendix \ref{proof_lem_var_q}. $\Box$.
\end{lemma}

\vspace{-0.15in}
\subsection{Optimal Quantile $q^*(\alpha)$}
\vspace{-0.05in}

We  choose $q = q^*(\alpha)$ so that the asymptotic variance (\ref{eqn_var_q}) is minimized, i.e.,\vspace{-0.15in}

{\small\begin{align}\label{eqn_g}
q^*(\alpha) = \underset{q}{\text{argmin}} \ g(q;\alpha), \  \ \ \ \  \  g(q;\alpha) = \frac{q-q^2}{f^2_X\left(W;\alpha,1\right) W^2 }.
\end{align}}\vspace{-0.1in}

The convexity of $g(q;\alpha)$ is important. Graphically, $g(q;\alpha)$ is a convex function of $q$, i.e., a unique minimum exists.  An algebraic proof, however, is difficult. Nevertheless, we can obtain analytical solutions when $\alpha = 1$ and $\alpha = 0+$.

\begin{lemma}\label{lem_convexity}
When $\alpha = 1$ or $\alpha = 0+$, the function $g(q;\alpha)$ defined in (\ref{eqn_g}) is a convex function of $q$. When $\alpha = 1$, the optimal $q^*(1) = 0.5$. When $\alpha =0+$, $q^*(0+)=0.203$ is the solution to $-\log q^* + 2q^* -2 = 0$.

\textbf{Proof:} See Appendix \ref{proof_lem_convexity}. $\Box$.
\end{lemma}

It is also easy to show that when $\alpha = 2$, $q^*(2) = 0.862$.

\begin{figure}[h]
\begin{center}
\mbox{\hspace{-0.in}
\subfigure[$q^*$]{\includegraphics[height = 1.5in]{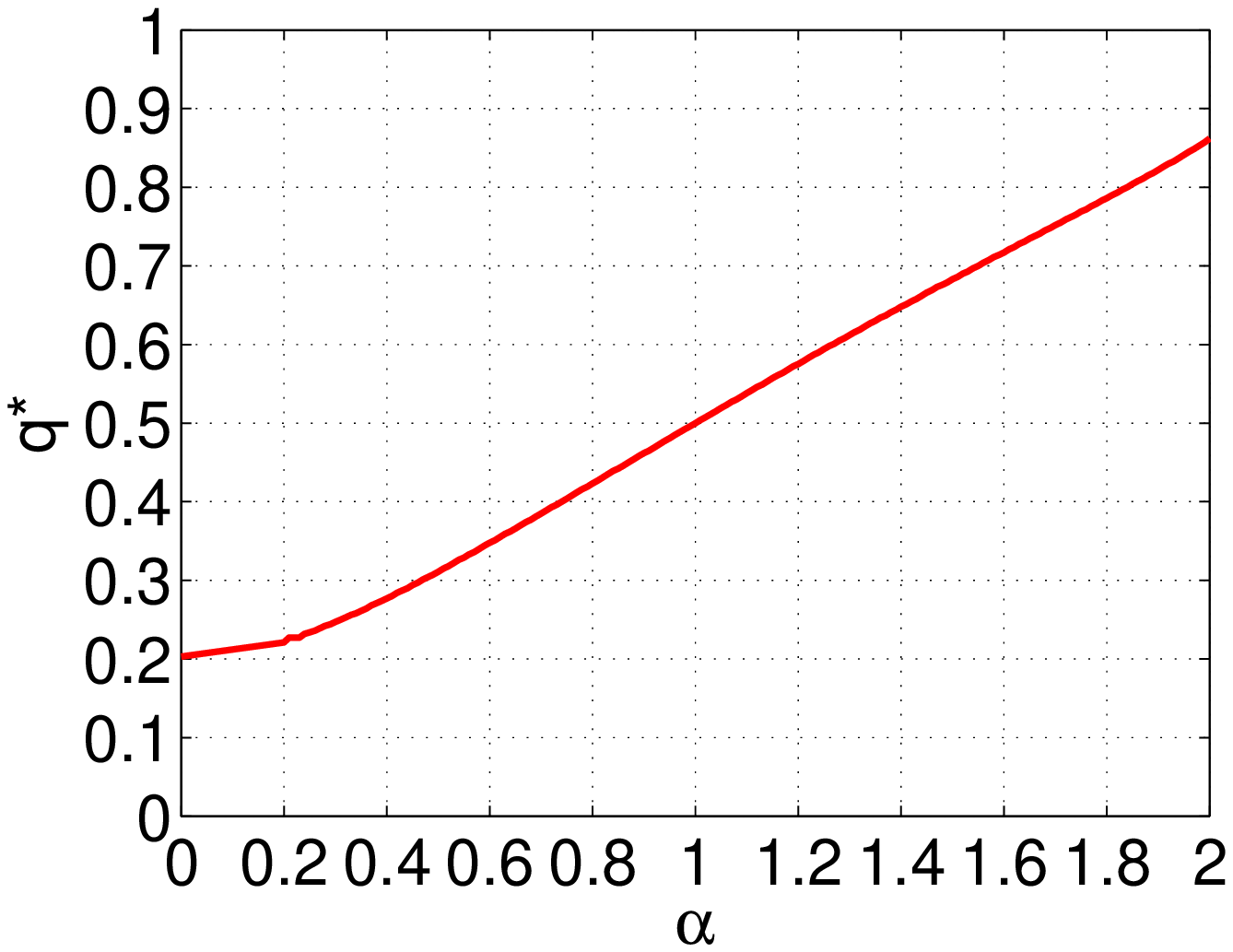}}\hspace{0.2in}
\subfigure[$W^\alpha(q^*)$]{\includegraphics[height = 1.5in]{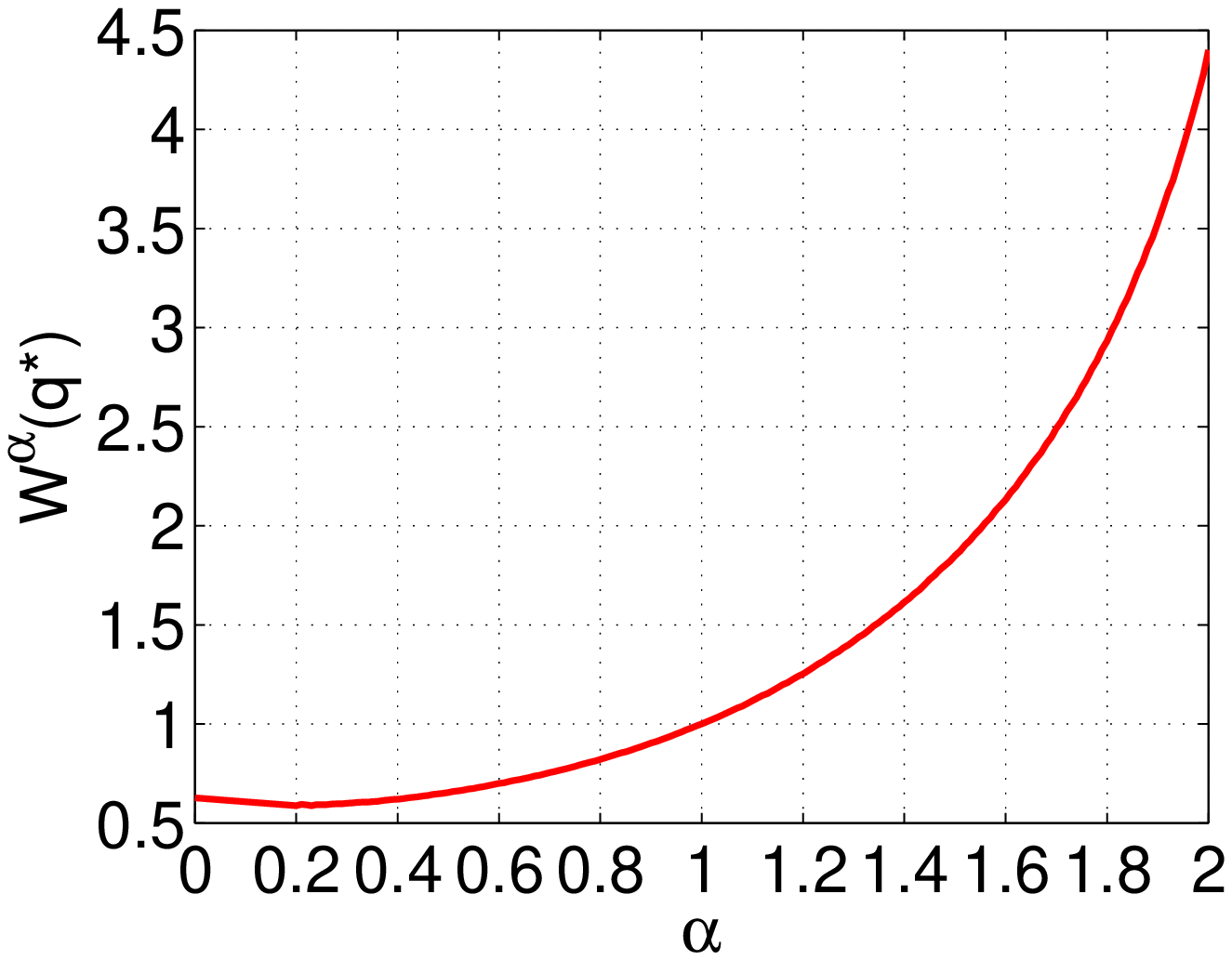}}
}
\end{center}
\vspace{-0.3in}
\caption{(a) The optimal values for $q^*(\alpha)$, which minimizes asymptotic variance of $\hat{d}_{(\alpha),q}$, i.e., the solution to (\ref{eqn_g}). (b) The constant $W^\alpha(q^*) = \{q^*\text{-quantile}\{|S(\alpha,1)|\}\}^\alpha$.
}\label{fig_opt_quantile}\vspace{-0.2in}
\end{figure}

We denote the {\em optimal quantile} estimator by $\hat{d}_{(\alpha),oq}$, which is same as $\hat{d}_{(\alpha),q^*}$.
For general $\alpha$, we resort to numerical solutions, as presented in Figure \ref{fig_opt_quantile}.

\vspace{-0.2in}

\vspace{-0.15in}
\subsection{Bias Correction}
\vspace{-0.05in}

Although $\hat{d}_{(\alpha),oq}$ (i.e., $\hat{d}_{(\alpha),q^*}$) is asymptotically (as $k\rightarrow\infty$) unbiased, it is seriously biased for small $k$. Thus, it is practically important to remove the bias. The unbiased version of the {\em optimal quantile} estimator is\vspace{-0.2in}

\begin{figure}[h]\vspace{-0.2in}
\begin{center}
\mbox{\hspace{-0.2in}
{\includegraphics[height = 1.5in]{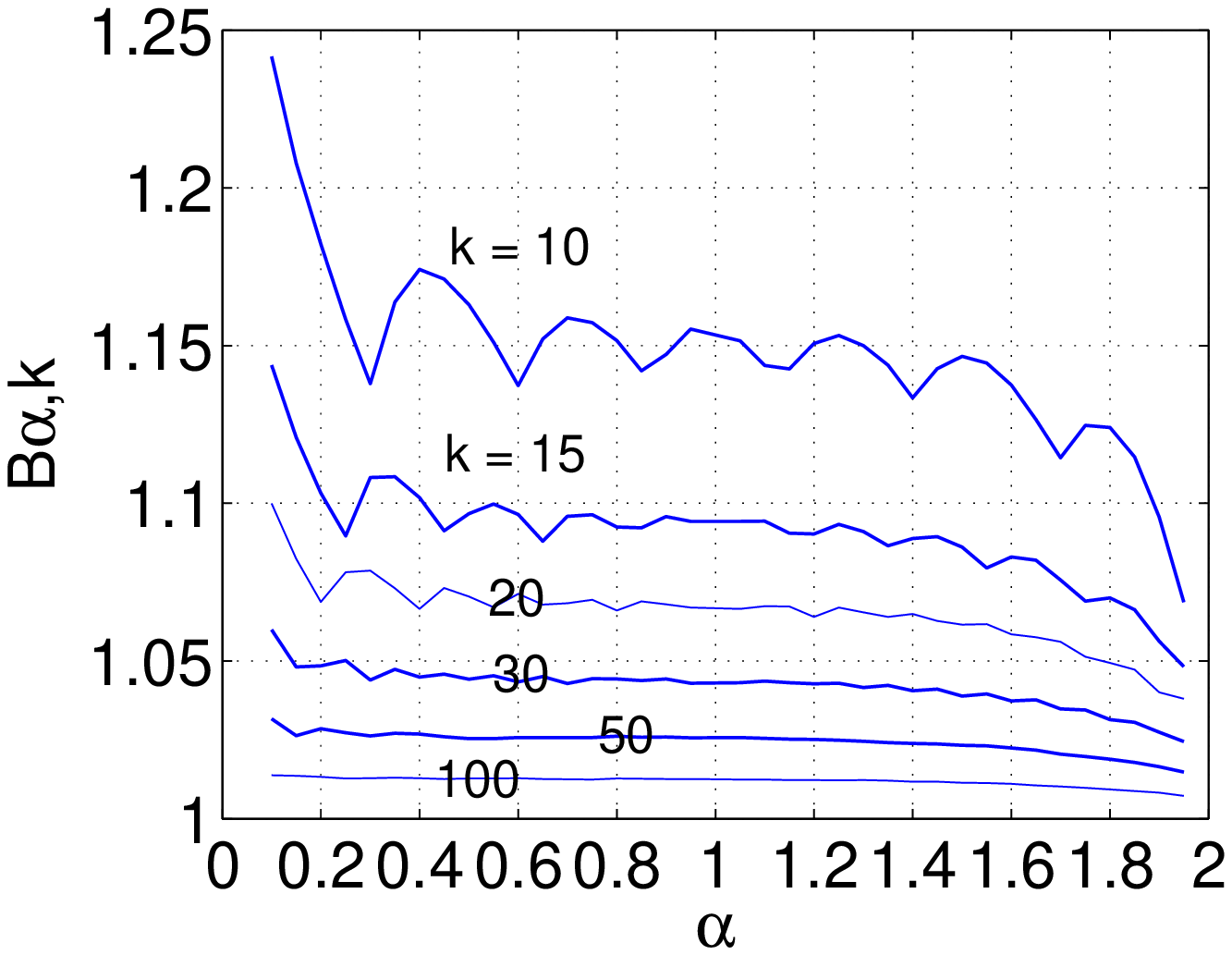}}\hspace{0.2in}
{\includegraphics[height = 1.5in]{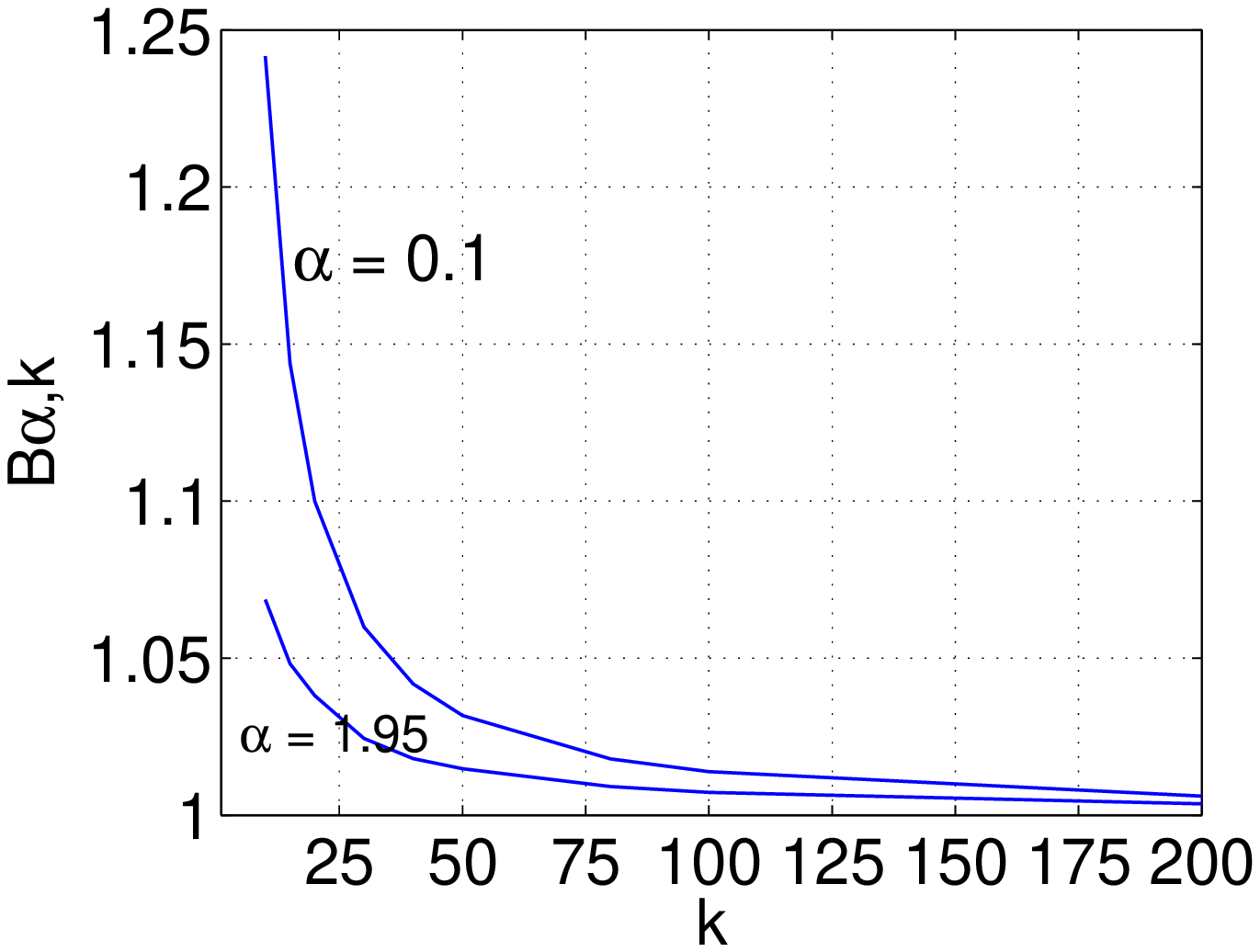}}
}
\end{center}
\vspace{-0.3in}
\caption{The bias correction factor $B_{\alpha,k}$ in (\ref{eqn_oqc}), obtained from $10^8$ simulations for every combination of $\alpha$ (spaced at 0.05) and $k$. {\small$B_{\alpha,k} = \text{E}\left(\hat{d}_{(\alpha),oq}; d_{(\alpha)}=1\right)$}.
}\label{fig_bias}
\end{figure}

{\small
\begin{align}\label{eqn_oqc}
\hat{d}_{(\alpha),oq,c} = \hat{d}_{(\alpha),oq}/B_{\alpha,k},
\end{align}}\vspace{-0.2in}

\noindent where $B_{\alpha,k}$ is the expectation of $\hat{d}_{(\alpha),oq}$ at $d_{(\alpha)} =1$. For $\alpha = 1$, $0+$, or $2$, we can evaluate the expectations  (i.e., integrals) analytically or by numerical integrations. For general $\alpha$, as the probability density is not available, the task is difficult and prone to numerical instability. On the other hand, since the Monte-Carlo simulation is a popular alternative for evaluating difficult integrals, a practical solution is to simulate the expectations, as presented in Figure \ref{fig_bias}.

Figure \ref{fig_bias} illustrates that $B_{\alpha,k}>1$, meaning that this correction also reduces  variance while removing bias (because {\small $\text{Var}(x/c)= \text{Var}(x)/c^2$}).  For example, when $\alpha = 0.1$ and $k = 10$, $B_{\alpha,k}\approx 1.24$, which is  significant, because $1.24^2 = 1.54$ implies a $54\%$ difference in terms of  variance, and  even more considerable in terms of the mean square errors  MSE = variance + bias$^2$.

$B_{\alpha,k}$ can be tabulated for small $k$, and absorbed into other coefficients, i.e., this does not increase the computational cost at run time. We fix $B_{\alpha,k}$ as reported in Figure \ref{fig_bias}.  The simulations in Section \ref{sec_simulations} directly used those fixed $B_{\alpha,k}$ values.

\vspace{-0.1in}
\subsection{Computational Efficiency}
\vspace{-0.05in}

\begin{figure}[h]
\begin{center}
\mbox{\hspace{-0.in}
{\includegraphics[height = 1.5in]{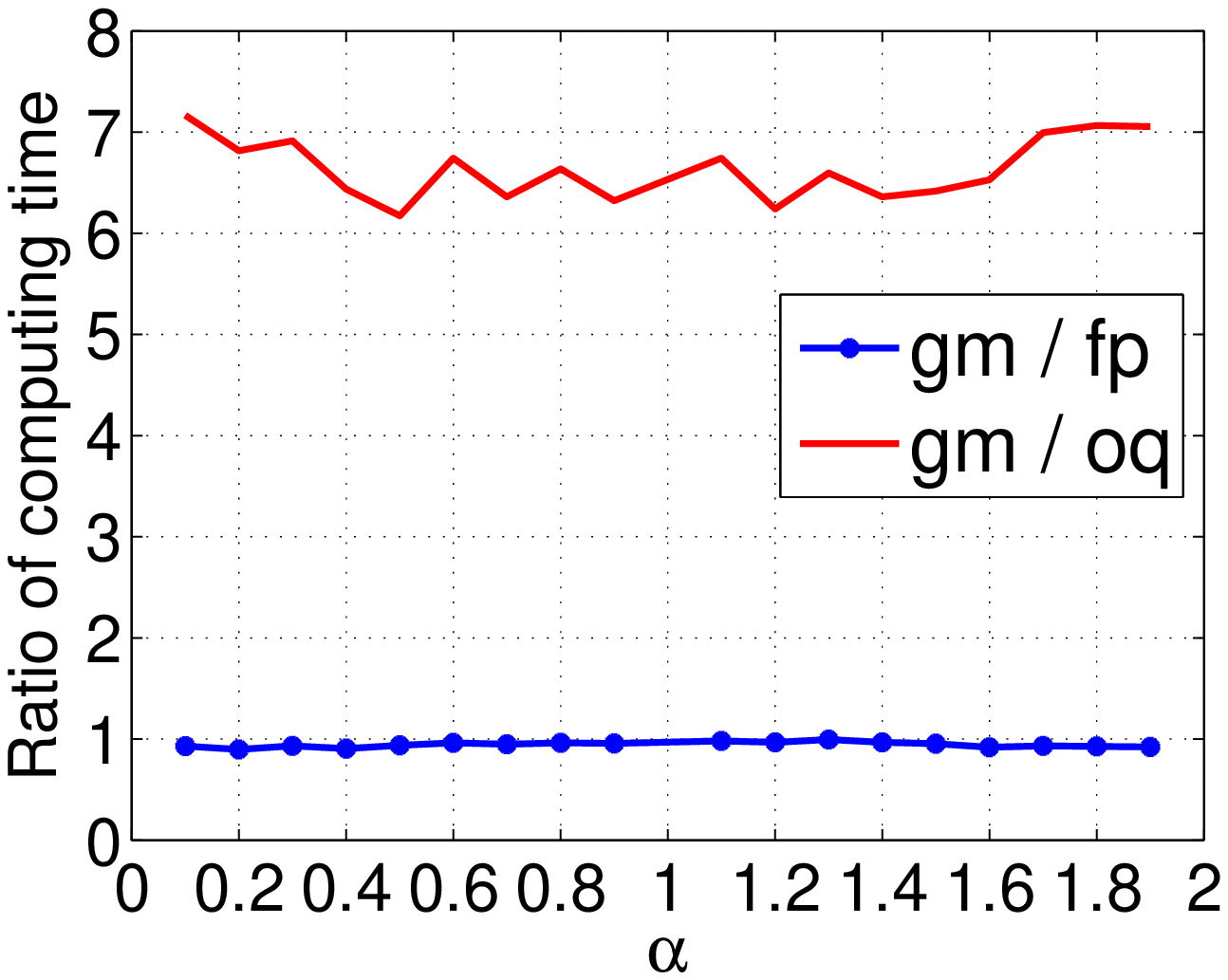}}\hspace{0.2in}
{\includegraphics[height = 1.5in]{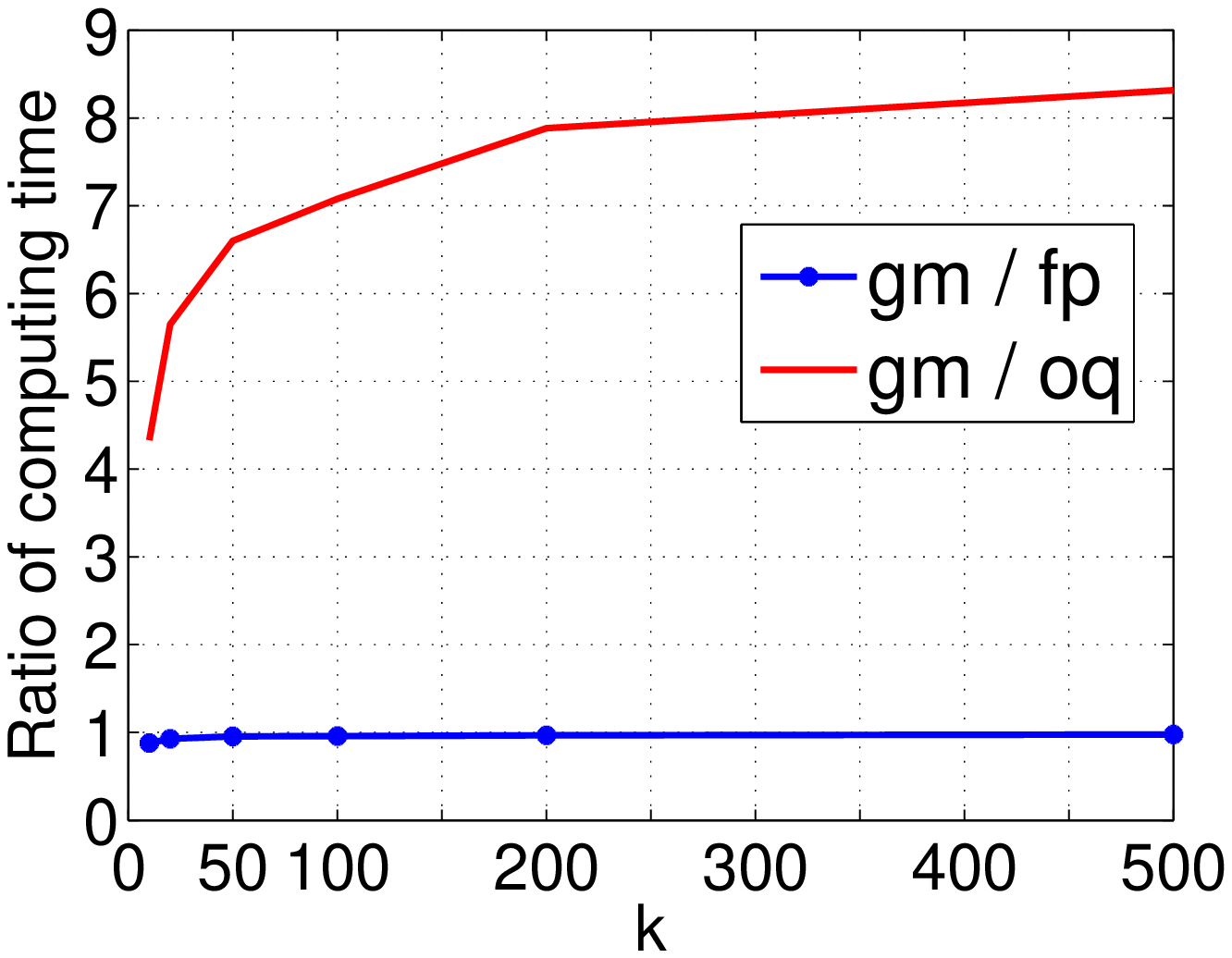}}
}
\end{center}
\vspace{-0.3in}
\caption{Relative computational cost ({\small$\hat{d}_{(\alpha),gm}$} over {\small$\hat{d}_{(\alpha),oq,c}$} and {\small$\hat{d}_{(\alpha),gm}$} over {\small$\hat{d}_{(\alpha),fp}$}), from $10^6$ simulations at each combination of $\alpha$ and $k$. The left panel averages over all $k$ and the right panel averages over all $\alpha$. Note  that the cost of {\small$\hat{d}_{(\alpha),oq,c}$} includes evaluating the $\alpha$th moment once. }\label{fig_compu_ratio}\vspace{-0.15in}
\end{figure}

Figure \ref{fig_compu_ratio} compares the computational costs of the {\em geometric mean}, the {\em fractional power}, and the {\em optimal quantile} estimators. The {\em harmonic mean} estimator was not included as it costs very similarly to the {\em fractional power} estimator.

We used the build-in function ``pow''in gcc for evaluating the fractional powers. We implemented a
``quick select'' algorithm, which is similar to quick sort and requires on average linear time. For simplicity, our implementation used recursions and the middle element as pivot. Also, to ensure fairness, for all estimators, coefficients which are functions of $\alpha$ and/or $k$ were pre-computed.

\vspace{-0.2in}

Normalized by the computing time of {\small$\hat{d}_{(\alpha),gm}$}, we observe that relative computational efficiency does not strongly depend on $\alpha$.  We do observe that the ratio of computing time of {\small$\hat{d}_{(\alpha),gm}$} over that of {\small$\hat{d}_{(\alpha),oq,c}$} increases consistently with increasing $k$. This is because in the definition of {\small$\hat{d}_{(\alpha),oq}$} (and hence also {\small$\hat{d}_{(\alpha),oq,c}$}), it is required to evaluate the fractional power once, which contributes to the total computing time more significantly at smaller $k$.

Figure \ref{fig_compu_ratio} illustrates that, (A) the {\em geometric mean} estimator and the {\em fractional power} estimator are similar in terms of computational efficiency; (B) the {\em optimal quantile} estimator is nearly one order of magnitude more computationally efficient than the {\em geometric mean}  and  {\em fractional power} estimators. Because we implemented a ``na\'ive'' version of ``quick select'' using recursions and simple pivoting, the actual improvement may be  more significant. Also, if applications require only {\small$d_{(\alpha)}^{1/\alpha}$}, then no fractional power operations are needed for  {\small$\hat{d}_{(\alpha),oq,c}$} and the improvement will be even more considerable.

\vspace{-0.1in}
\subsection{Error (Tail) Bounds}

Error (tail) bounds are essential for determining $k$. The variance alone is not sufficient for that purpose. If an estimator of $d$, say $\hat{d}$, is normally distributed, {\small$\hat{d} \sim N\left(d, \frac{1}{k}V\right)$},  the variance suffices for choosing $k$ because its error (tail) probability {\small $\mathbf{Pr}\left(|\hat{d}-d|\geq \epsilon d\right) \leq 2\exp\left(-k\frac{\epsilon^2}{2V}\right)$} is determined by $V$. In general, a reasonable estimator will be asymptotically normal, for small enough $\epsilon$ and large enough $k$. For a finite $k$ and a fixed $\epsilon$, however, the normal approximation may be (very) poor. This is especially true for the {\em fractional power} estimator, {\small$\hat{d}_{(\alpha),fp}$.}

Thus, for a good motivation, Lemma \ref{lem_bounds} provides the error (tail) probability bounds of $\hat{d}_{(\alpha),q}$ for any $q$, not just the optimal quantile $q^*$.

\begin{lemma}\label{lem_bounds}
Denote {\small$X \sim S(\alpha,d_{(\alpha)})$} and its probability density function by {\small$f_X(x;\alpha,d_{(\alpha)})$} and cumulative function by {\small$F_X(x;\alpha,d_{(\alpha)})$}. Given {\small$x_j\sim S(\alpha,d_{(\alpha)})$}, i.i.d., $j = 1$ to $k$. Using {\small$\hat{d}_{(\alpha),q}$} in (\ref{eqn_quantile}), then\vspace{-0.15in}

{\small\begin{align}
&\mathbf{Pr}\left( \hat{d}_{(\alpha),q} \geq(1+\epsilon)  d_{(\alpha)}\right)
\leq \exp\left(-k\frac{\epsilon^2}{G_{R,q}}\right), \hspace{0.in} \epsilon >0, \\
&\mathbf{Pr}\left( \hat{d}_{(\alpha),q} \leq(1-\epsilon)  d_{(\alpha)}\right)
\leq \exp\left(-k\frac{\epsilon^2}{G_{L,q}}\right), \hspace{0.in} 0< \epsilon <1,
\end{align}}\vspace{-0.4in}

{\small\begin{align}\label{eqn_G_R}
&\frac{\epsilon^2}{G_{R,q}} = -(1-q)\log\left(2-2F_R\right) - q \log(2F_R-1) + (1-q)\log (1-q) + q\log q, \\\label{eqn_G_L}
&\frac{\epsilon^2}{G_{L,q}} = -(1-q)\log\left(2-2F_L\right)- q \log (2F_L-1)+ (1-q)\log (1-q) + q\log q,\\\notag\vspace{0.1in}
&W  = F_X^{-1}((q+1)/2;\alpha,1) = q\text{-quantile}\{|S(\alpha,1)|\},\\\notag
&F_R = F_X\left((1+\epsilon)^{1/\alpha}W;\alpha,1\right), \hspace{0.2in} F_L = F_X\left((1-\epsilon)^{1/\alpha}W;\alpha,1\right).
\end{align}}\vspace{-0.15in}

As $\epsilon\rightarrow 0+$\vspace{-0.1in}
{\small\begin{align}\label{eqn_G_RL_limit}
&\underset{\epsilon\rightarrow0+}{\lim} G_{R,q} =
\underset{\epsilon\rightarrow0+}{\lim} G_{L,q} = \frac{q(1-q)\alpha^2/2}{f_X^2\left(W;\alpha,1\right)W^2}.
\end{align}}
\textbf{Proof:} \ \ See Appendix \ref{proof_lem_bounds}. $\Box$
\end{lemma}

The limit in (\ref{eqn_G_RL_limit}) as $\epsilon\rightarrow 0$ is precisely twice the asymptotic variance factor of $\hat{d}_{(\alpha),q}$ in (\ref{eqn_var_q}), consistent with the normality approximation mentioned previously. This  explains why we  express the constants as {\small$\epsilon^2/G$}. (\ref{eqn_G_RL_limit}) also indicates that the tail bounds achieve the ``optimal rate'' for this estimator, in the language of large deviation theory.

By the Bonferroni  bound, it is easy to determine the sample size $k$
{\small\begin{align}\notag\vspace{-0.15in}
&\mathbf{Pr}\left(|\hat{d}_{(\alpha),q} - d_{(\alpha)}|\geq \epsilon d_{(\alpha)}\right) \leq 2 \exp\left(-k\frac{\epsilon^2}{G}\right) \leq \delta/(n^2/2)
\Longrightarrow k \geq \frac{G}{\epsilon^2}\left(2\log n - \log \delta\right).
\end{align}} \vspace{-0.2in}

\begin{lemma}\label{lem_JL_quantile}
Using  $\hat{d}_{(\alpha),q}$ with {\small$k\geq \frac{G}{\epsilon^2}\left(2\log n - \log \delta\right)$}, any pairwise $l_\alpha$ distance among $n$  points can be approximated within a $1\pm\epsilon$ factor with probability $\geq1-\delta$. It suffices to let {\small$G = \max\{G_{R, q}, G_{L, q}\}$}, where {\small$G_{R, q}$, $G_{L, q}$} are defined in Lemma \ref{lem_bounds}.
\end{lemma}

The Bonferroni bound can be unnecessarily conservative. It is often reasonable to replace {\small$\delta/(n^2/2)$} by {\small$\delta/T$}, meaning that except for a $1/T$ fraction of pairs, any  distance can be approximated within a $1\pm\epsilon$ factor with probability $1-\delta$.

\begin{figure}[h]\vspace{-0.2in}
\begin{center}
\mbox{\hspace{-0.in}
{\includegraphics[height = 1.5in]{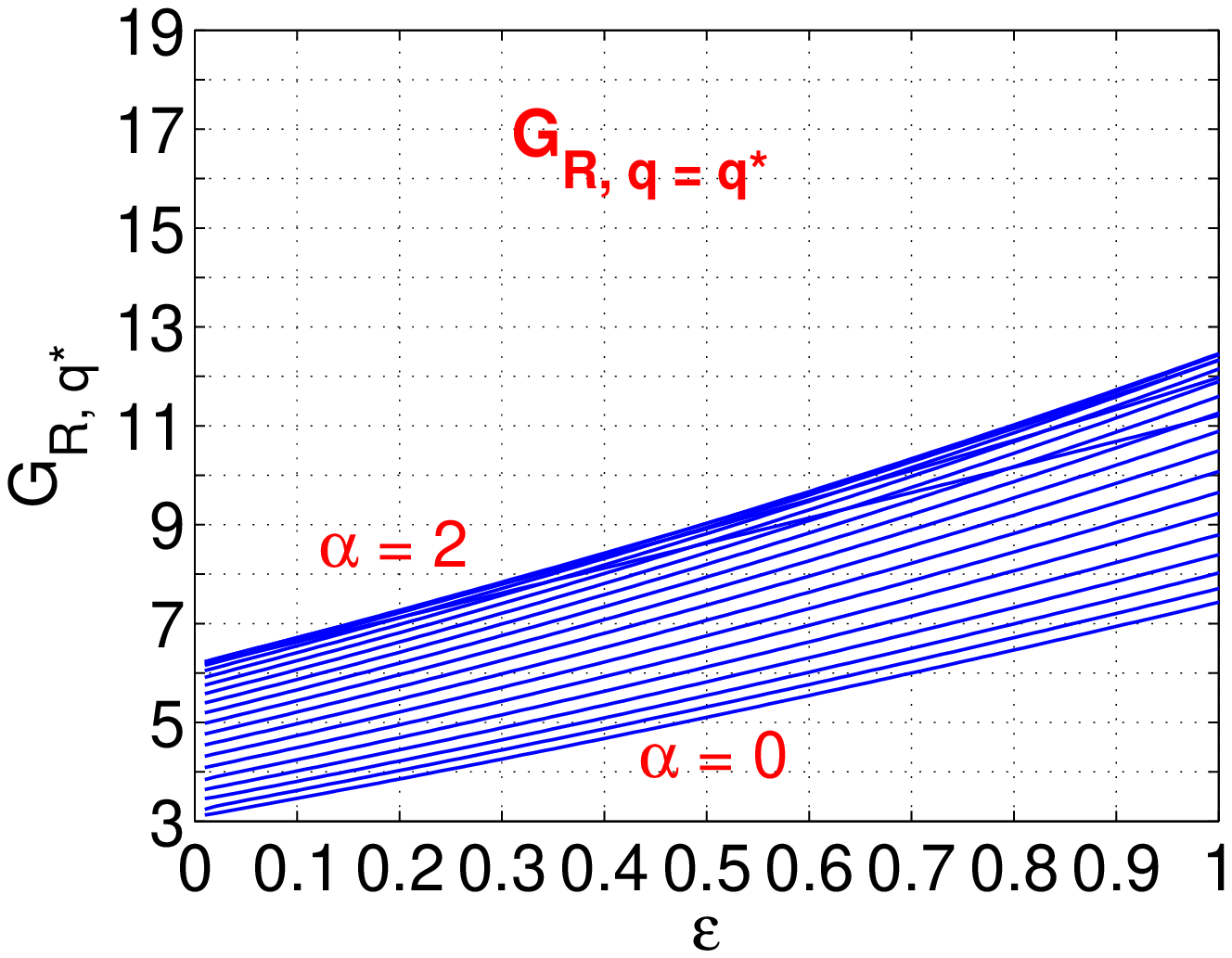}}\hspace{0.2in}
{\includegraphics[height = 1.5in]{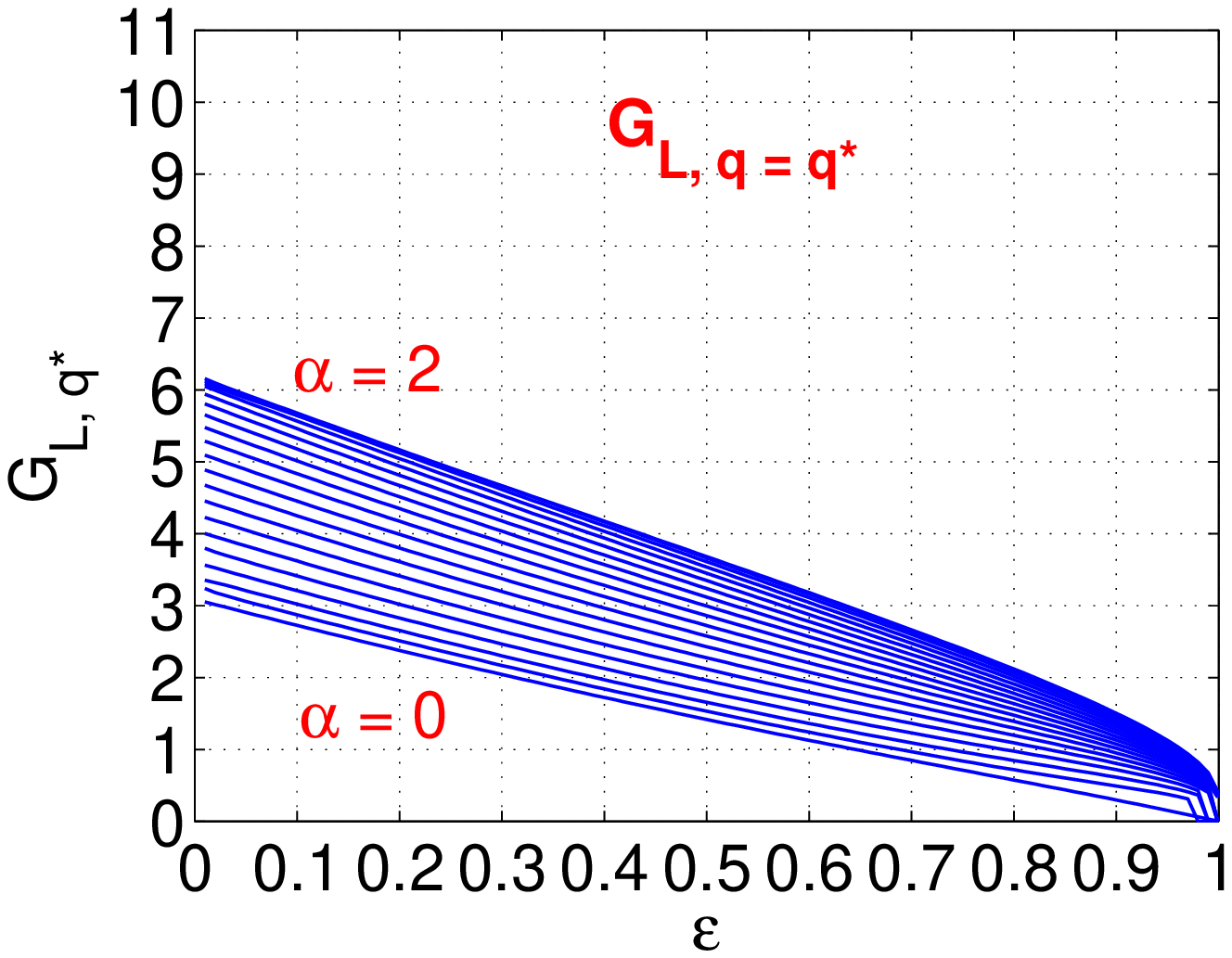}}
}\vspace{-0.1in}
\mbox{\hspace{-0.in}
{\includegraphics[height = 1.5in]{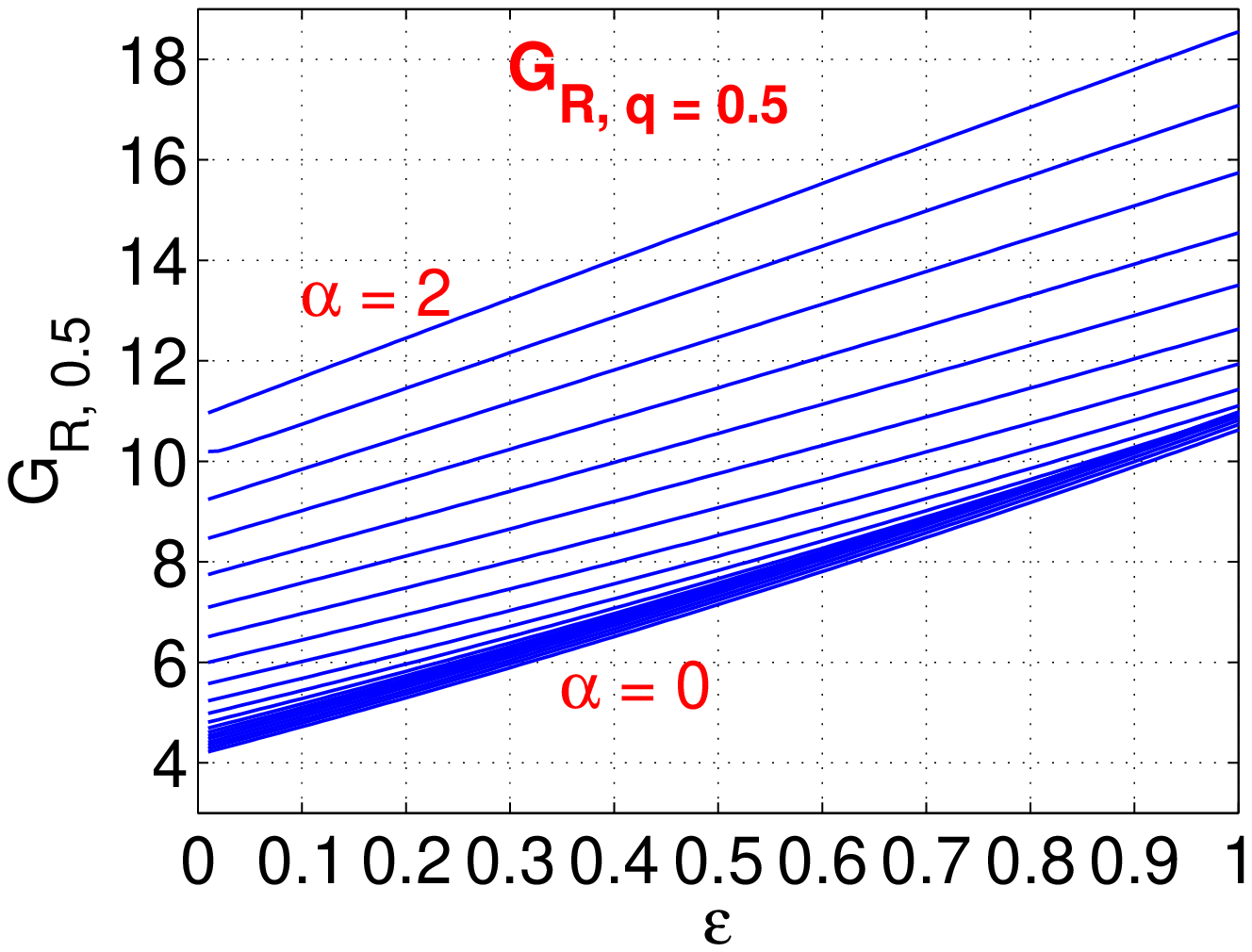}}\hspace{0.2in}
{\includegraphics[height = 1.5in]{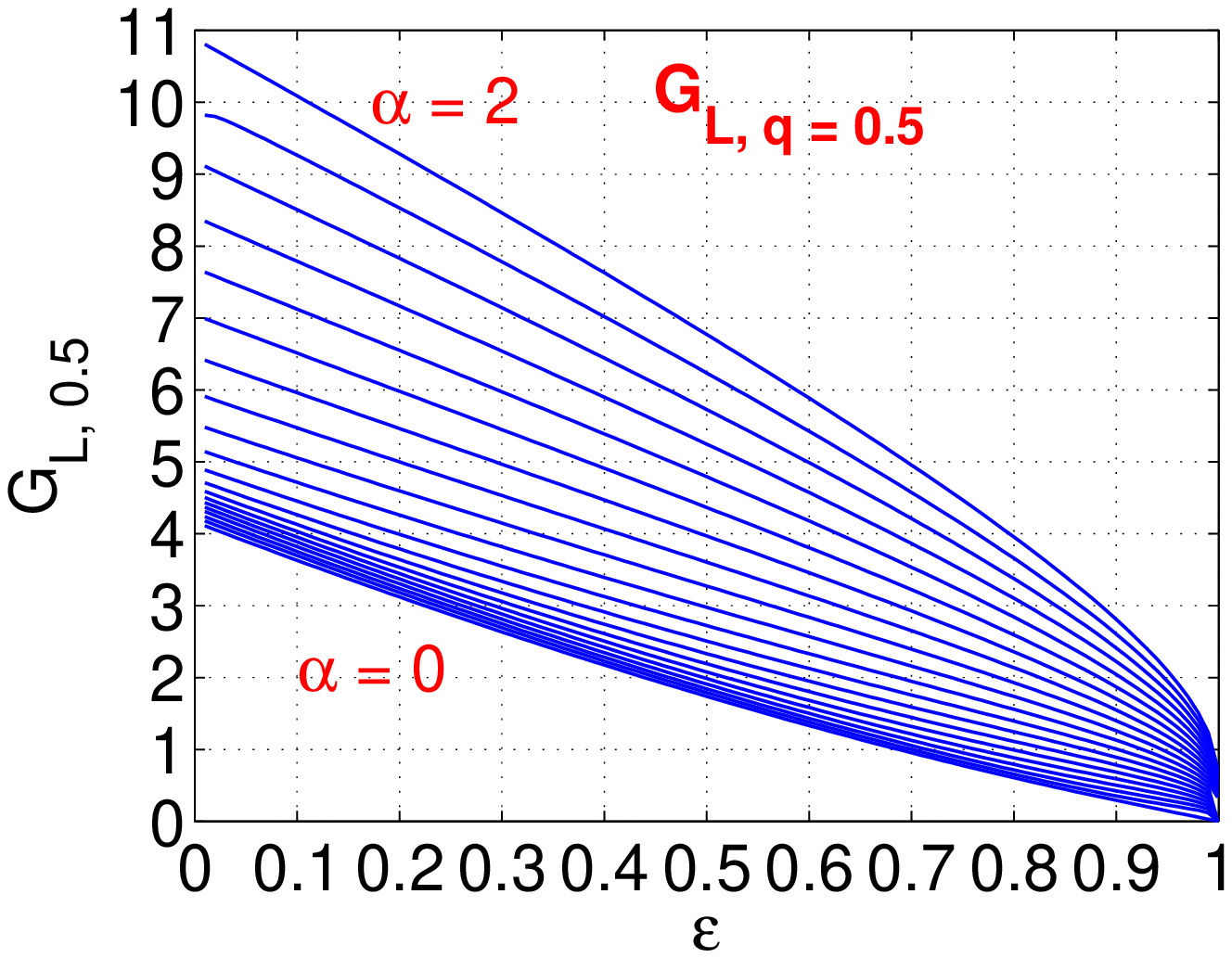}}
}
\end{center}
\vspace{-0.3in}
\caption{Tail bound constants for quantile estimators; the lower the better. Upper panels: optimal quantile estimators $\hat{d}_{(\alpha),q^*}$. Lower panels: median estimators $\hat{d}_{(\alpha),q=0.5}$.   }\label{fig_bounds}\vspace{-0.2in}
\end{figure}

Figure \ref{fig_bounds} plots the error bound constants for $\epsilon<1$, for both the recommended {\em optimal quantile} estimator $\hat{d}_{(\alpha),oq}$ and the baseline {\em sample median} estimator {\small$\hat{d}_{(\alpha),q=0.5}$}. Although we choose {\small$\hat{d}_{(\alpha),oq}$} based on the asymptotic variance, it turns out {\small$\hat{d}_{(\alpha),oq}$} also exhibits (much) better tail behaviors (i.e., smaller constants) than {\small$\hat{d}_{(\alpha),q=0.5}$}, at least in the range of $\epsilon<1$.

Consider {\small$k = \frac{G}{\epsilon^2}\left(\log 2T - \log \delta\right)$} (recall we suggest replacing $n^2/2$ by $T$), with $\delta = 0.05$, $\epsilon = 0.5$, and $T = 10$. Because $G_{R,q^*}\approx 5\sim 9$ around $\epsilon=0.5$, we obtain  $k \approx 120\sim215$, which is still a relatively large number (although the original dimension $D$ might be $10^6$). If we choose $\epsilon = 1$, then approximately $k \approx 40\sim 65$.

It is possible $k = 120\sim 215$ might be still conservative, for three reasons: (A) the tail bounds, although ``sharp," are still upper bounds; (B) using $G = \max\{G_{R,q^*}, G_{L,q^*}\}$ is conservative because $G_{L,q^*}$ is usually much smaller than $G_{R,q^*}$; (C) this type of tail bounds is based on relative error, which may be stringent  for small ($\approx 0$) distances.

In fact, some earlier studies on {\em normal random projections} (i.e., $\alpha =2$) \cite{Proc:Bingham_kdd01,Proc:Fradkin_KDD03}  empirically demonstrated that $k\geq50$ appeared sufficient.

\vspace{-0.15in}
\section{Simulations}\label{sec_simulations}
\vspace{-0.05in}

We resort to simulations for comparing the finite sample variances of various estimators and assessing the more precise error (tail) probabilities.

 One advantage of {\em stable random projections} is that  we know the (manually generated) distributions and the only source of errors is from the random number generations. Thus, we can simply rely on simulations to evaluate the estimators without using real data. In fact, after projections, the projected data follow exactly the stable distribution, regardless of the original real data distribution.

Without loss of generality, we simulate samples from $S(\alpha, 1)$ and estimate the scale parameter (i.e., 1) from the samples. Repeating the procedure $10^7$ times, we can reliably evaluate the mean square errors (MSE) and tail probabilities.

\vspace{-0.15in}
\subsection{Mean Square Errors (MSE)}
\begin{figure}[h]
\begin{center}
\mbox{\hspace{-0.in}
{\includegraphics[height = 1.5in]{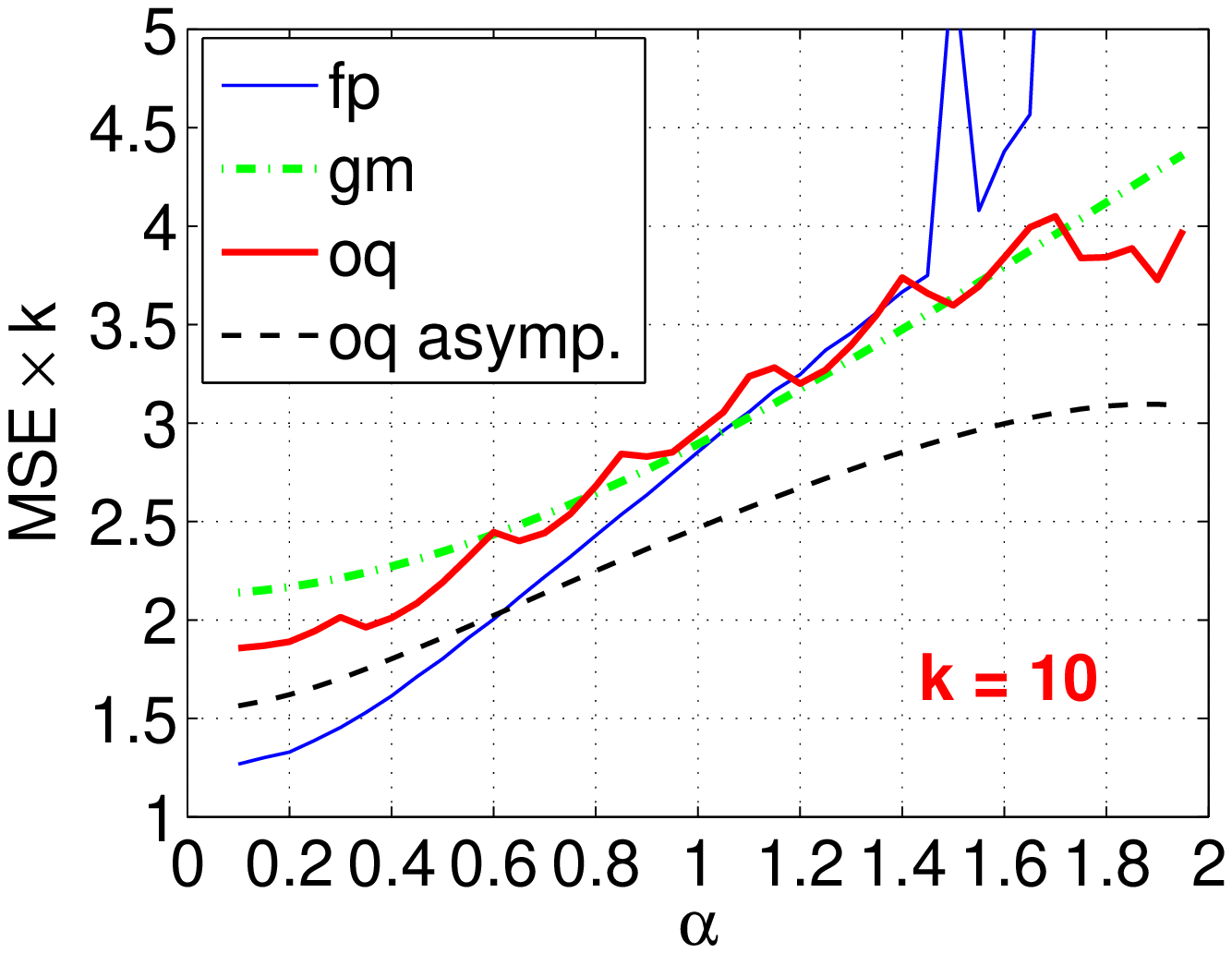}}\hspace{0.2in}
{\includegraphics[height = 1.5in]{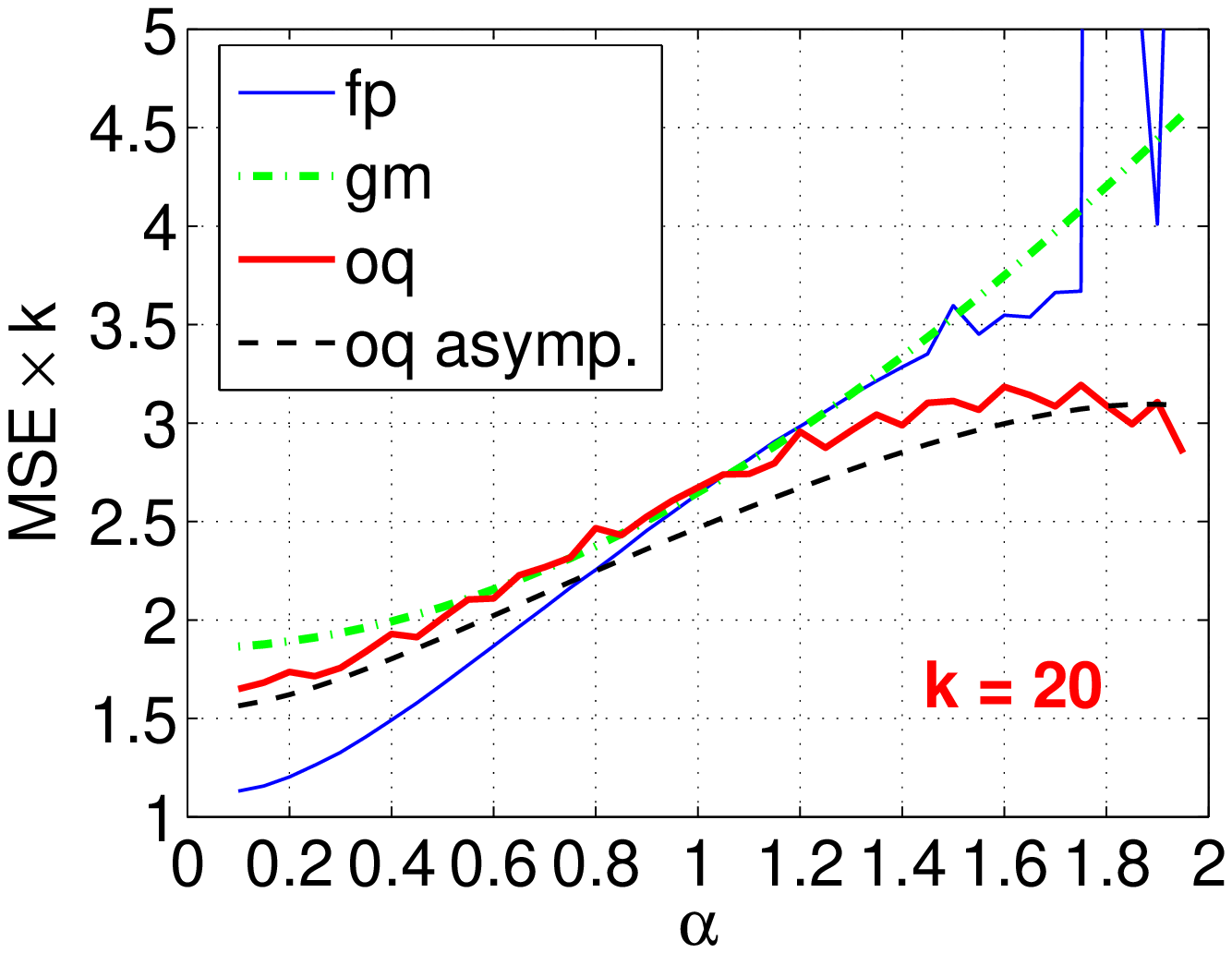}}
}\vspace{-0.1in}
\mbox{\hspace{-0.in}
{\includegraphics[height = 1.5in]{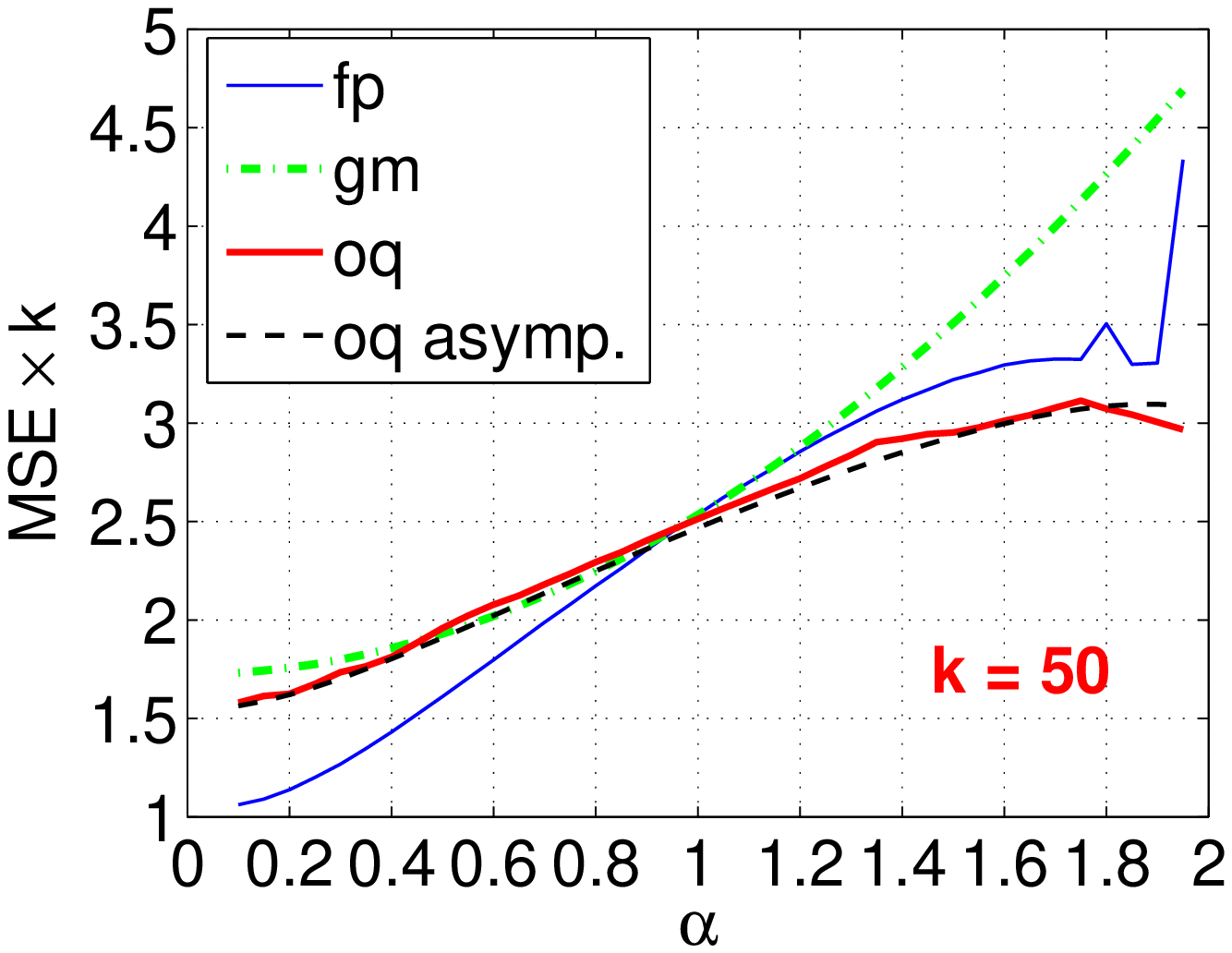}}\hspace{0.2in}
{\includegraphics[height = 1.5in]{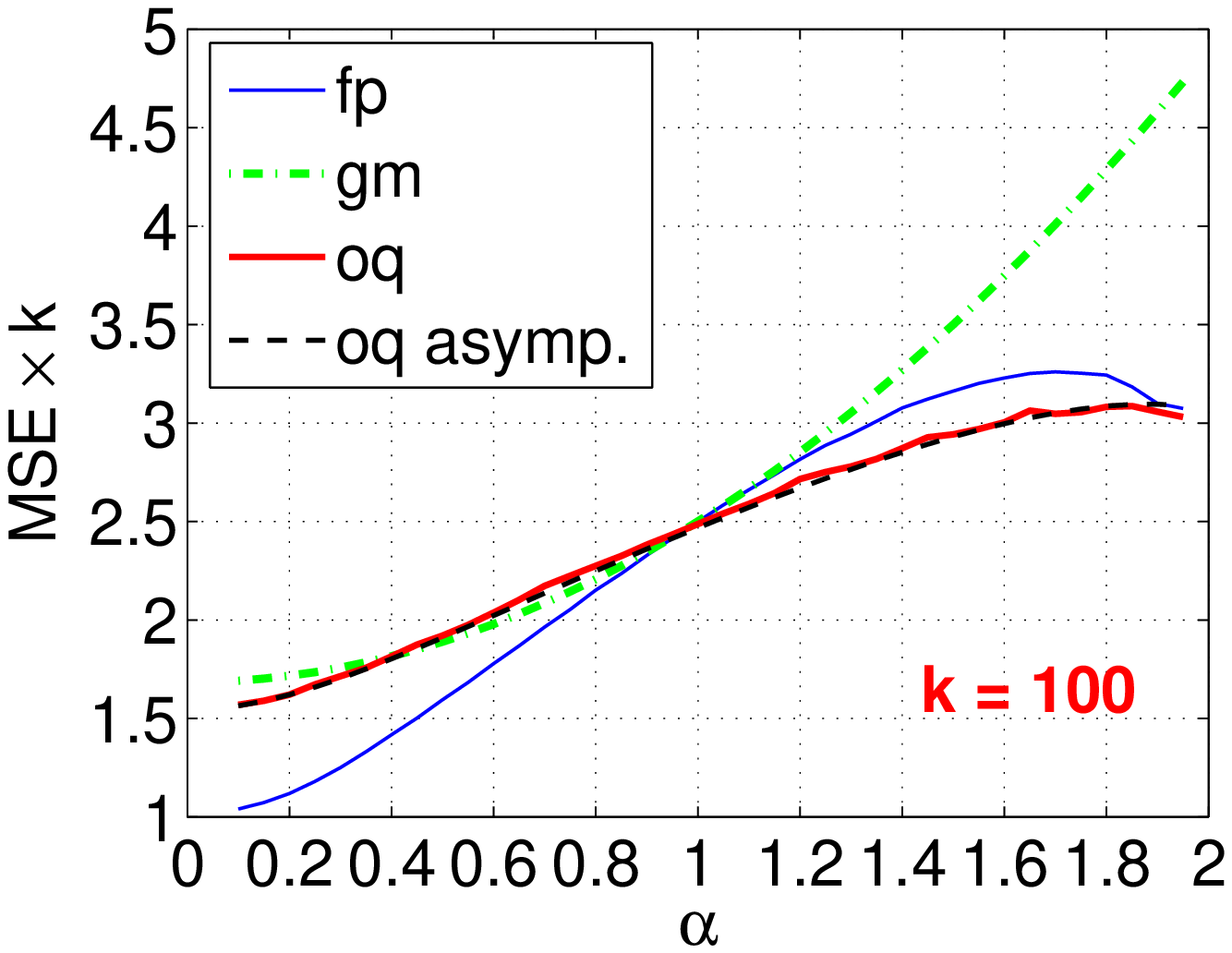}}
}
\end{center}
\vspace{-0.3in}
\caption{Empirical mean square errors (MSE, the lower the better), from $10^7$ simulations at every combination of $\alpha$ and $k$. The values are multiplied by $k$ so that four plots can be at about the same scale.  The MSE for the {\em geometric mean} (gm) estimator is computed exactly since closed-form expression exists. The lower dashed curves are the asymptotic variances of the {\em optimal quantile} (oq) estimator.
}\label{fig_simu_mse}\vspace{-0.2in}
\end{figure}

\vspace{-0.05in}
As illustrated in Figure \ref{fig_simu_mse}, in terms of the MSE,  the {\em optimal quantile} estimator $\hat{d}_{(\alpha),oq,c}$ outperforms both the {\em geometric mean} and {\em fractional power} estimators when $\alpha >1$ and $k\geq 20$. The {\em fractional power} estimator does not appear to be very suitable for $\alpha >1$, especially for $\alpha$ close to 2, even when the sample size $k$ is not too small (e.g., $k = 50$).  For $\alpha <1$, however, the {\em fractional power} estimator has good performance in terms of MSE, even for small $k$.

\vspace{-0.2in}
\subsection{Error(Tail) Probabilities}
\begin{figure}[h]
\begin{center}
\mbox{
{\includegraphics[height = 1.5in]{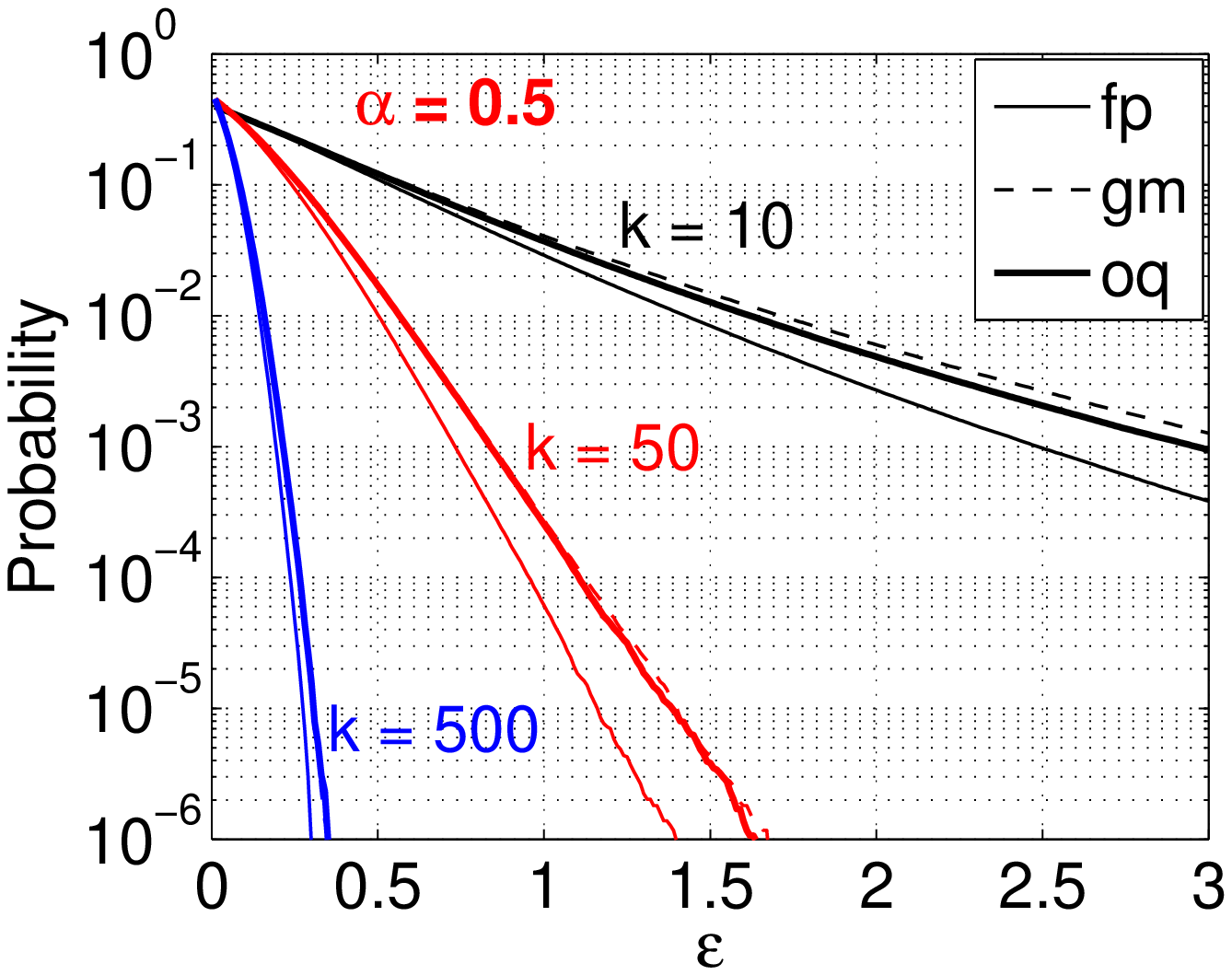}}\hspace{0.2in}
{\includegraphics[height = 1.5in]{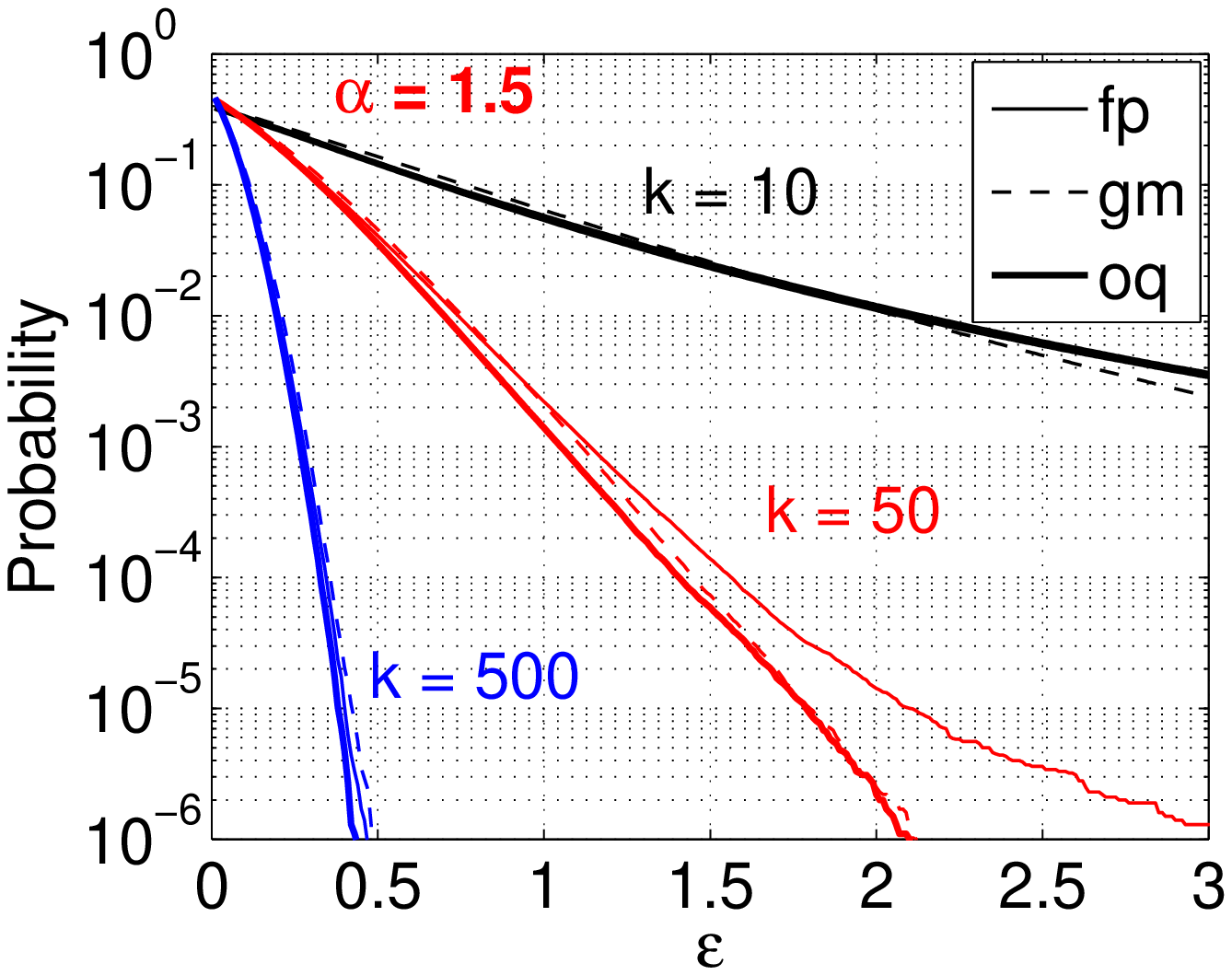}}
}\vspace{-0.1in}
\mbox{\hspace{-0.1in}
{\includegraphics[height = 1.5in]{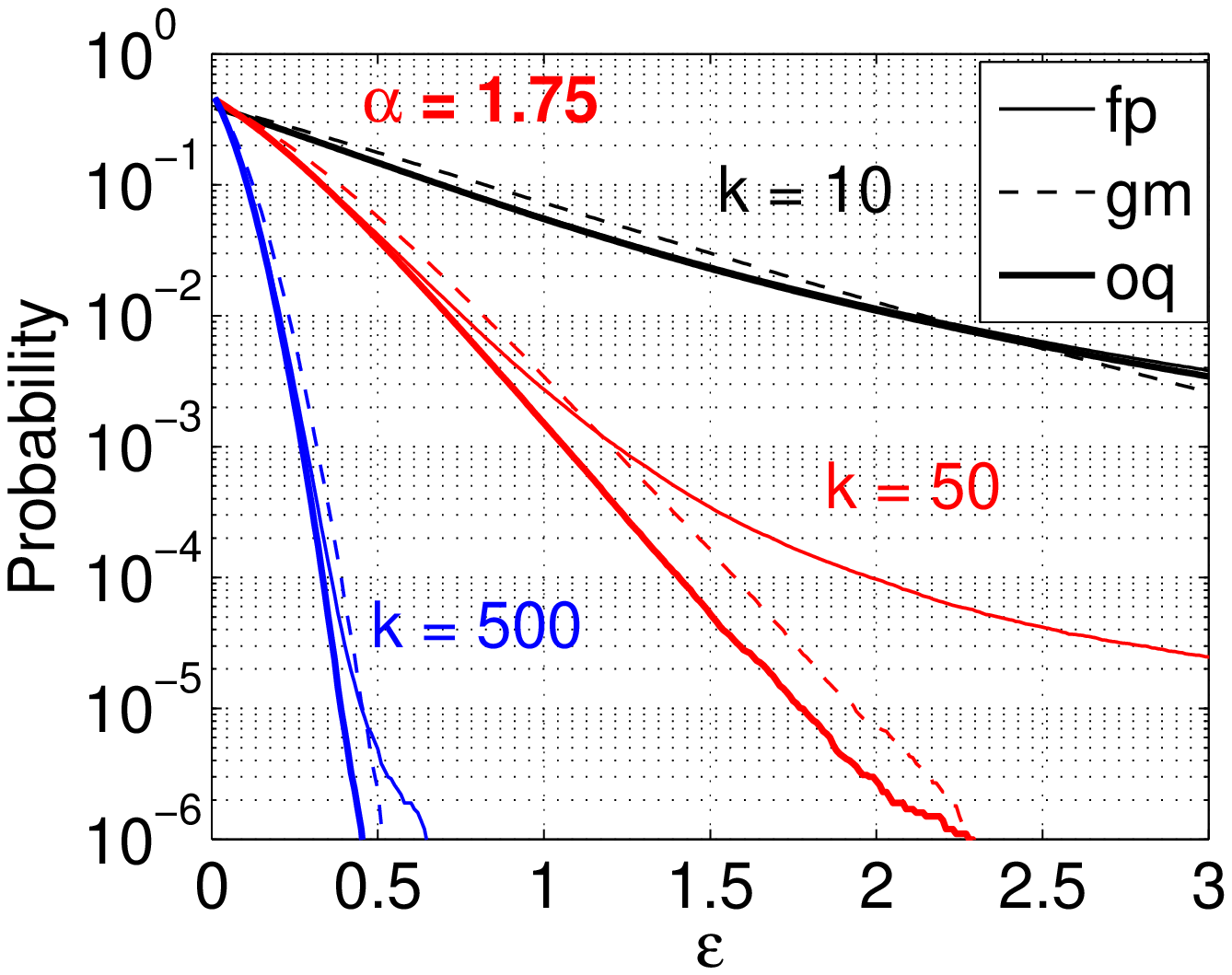}}\hspace{0.2in}
{\includegraphics[height = 1.5in]{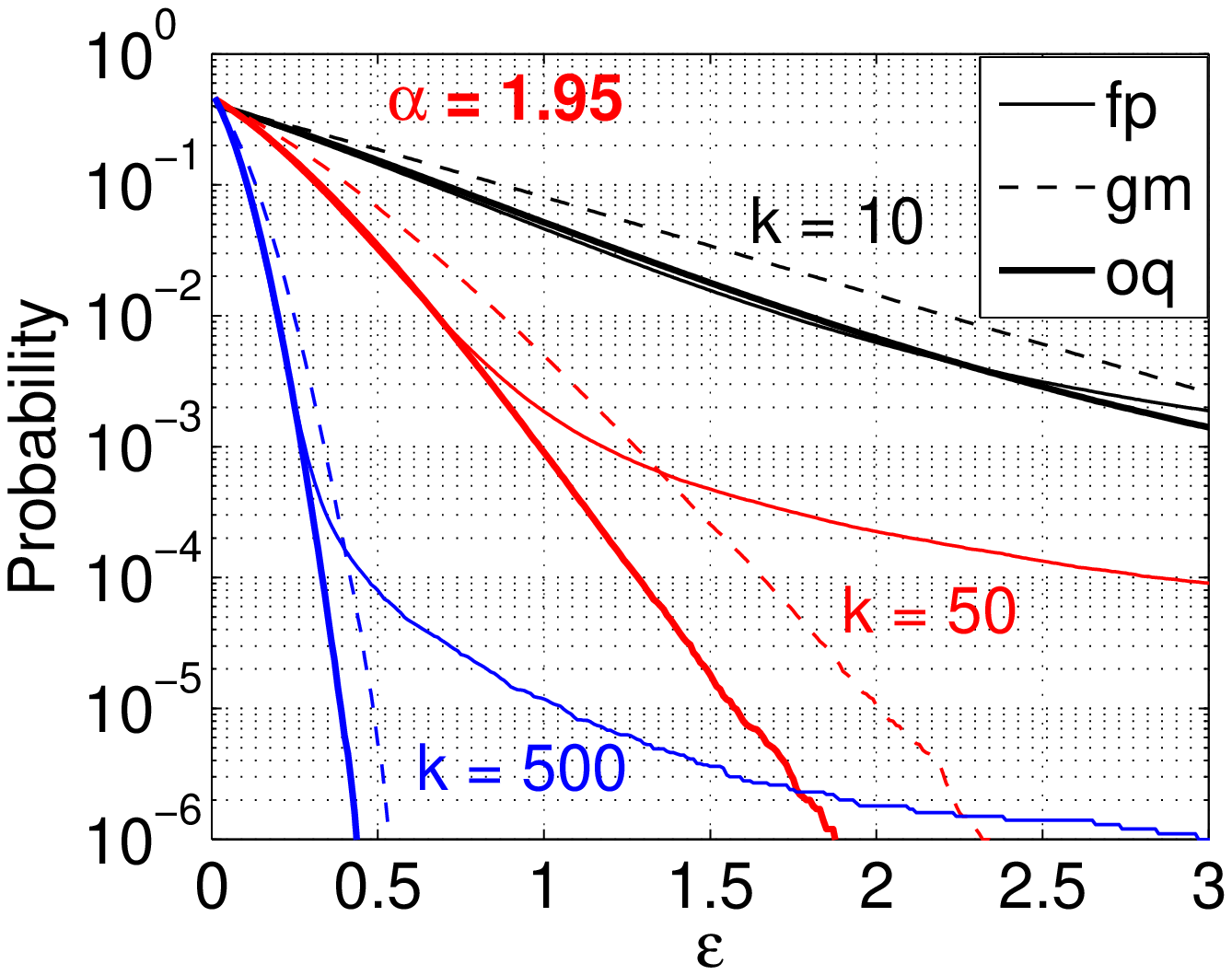}}
}
\end{center}
\vspace{-0.3in}
\caption{The right tail probabilities (the lower the better), from $10^7$ simulations at each combination of $\alpha$ and $k$.  }\label{fig_simu_tail}\vspace{-0.2in}
\end{figure}

\vspace{-0.05in}
Figure \ref{fig_simu_tail} presents the simulated right tail probabilities, {\small$\mathbf{Pr}\left(\hat{d}_{(\alpha)} \geq (1+\epsilon)d_{(\alpha)}\right)$}, illustrating that when $\alpha>1$, the {\em fractional power} estimator can exhibit very bad tail behaviors. For $\alpha<1$, the {\em fractional power} estimator demonstrates good performance at least for the probability range in the simulations.

Thus, Figure \ref{fig_simu_tail} demonstrates that the {\em optimal quantile} estimator consistently outperforms the {\em fractional power} and the {\em geometric mean} estimators when $\alpha >1$.

\vspace{-0.1in}
\section{The Related Work}
\vspace{-0.05in}

There have been many studies of {\em normal random projections} in machine learning, for dimension reduction in the $l_2$ norm, e.g., \cite{Proc:Fradkin_KDD03}, highlighted by the  Johnson-Lindenstrauss (JL) Lemma \cite{Article:JL84}, which says $k = O\left(\log n/\epsilon^2\right)$ suffices when using normal (or normal-like, e.g., \cite{Article:Achlioptas_JCSS03}) projection methods.

The method of {\em stable random projections}  is applicable for computing the $l_\alpha$ distances ($0<\alpha \leq 2$), not just for $l_2$. \cite[Lemma 1, Lemma 2, Theorem 3]{Article:Indyk_JACM06} suggested the {\em median} (i.e., $q = 0.5$ quantile) estimator for $\alpha =1$ and argued that the sample complexity bound should be $O\left(1/\epsilon^2\right)$ ($n = 1$ in their study). Their bound was not provided in an explicit form and required an ``$\epsilon$ is small enough'' argument. For $\alpha \neq 1$, \cite[Lemma 4]{Article:Indyk_JACM06} only provided a conceptual algorithm, which ``is not uniform.''
In this study, we prove the bounds for any $q$-quantile and any $0<\alpha\leq 2$ (not just $\alpha=1$), in explicit exponential forms, with no unknown constants and no restriction that ``$\epsilon$ is small enough.''

The quantile estimator for stable distributions was  proposed in statistics quite some time ago, e.g., \cite{Article:Fama_71,Article:McCulloch_86}. \cite{Article:Fama_71} mainly focused on $1\leq \alpha \leq 2$ and recommended using $q=0.44$ quantiles (mainly for the sake of smaller bias). \cite{Article:McCulloch_86} focused on $0.6\leq \alpha \leq 2$ and recommended $q=0.5$ quantiles.

This study considers all $0<\alpha \leq 2$ and recommends $q$ based on the minimum asymptotic variance. Because the bias can be easily removed (at least in the practical sense), it appears not necessary to use other quantiles only for the sake of smaller bias. Tail bounds, which are useful for choosing $q$ and $k$ based on confidence intervals, were not available in \cite{Article:Fama_71,Article:McCulloch_86}.

Finally, one might ask if there might be  better estimators.  For $\alpha = 1$, \cite{Article:Chernoff_67} proposed using a linear combination of quantiles (with carefully chosen coefficients) to obtain an asymptotically optimal estimator for the Cauchy scale parameter. While it is  possible to extend their result to general $0<\alpha<2$ (requiring some non-trivial work), whether or not it will be practically better than the {\em optimal quantile} estimator is unclear because the extreme quantiles severely affect the tail probabilities and finite-sample variances and hence some kind of truncation (i.e., discarding some samples at extreme quantiles) is necessary. Also, exponential tail bounds of the linear combination of quantiles may not exist or may not be feasible to derive. In addition, the {\em optimal quantile} estimator is computationally more efficient.

\vspace{-0.15in}
\section{Conclusion}
\vspace{-0.05in}

Many machine learning algorithms operate on the training data only through pairwise distances. Computing, storing, updating and retrieving the ``matrix'' of pairwise distances is challenging in applications involving massive, high-dimensional, and possibly streaming, data. For example, the pairwise distance matrix can not fit in memory when the number of observations exceeds $10^6$ (or even $10^5$).

The method of {\em stable random projections} provides an efficient mechanism for computing pairwise distances using low memory, by transforming the original high-dimensional data into {\em sketches}, i.e., a small number of samples from $\alpha$-stable distributions, which are much easier to store and retrieve.

This method provides a uniform scheme for computing the $l_\alpha$ pairwise distances for all $0<\alpha\leq 2$. Choosing an appropriate $\alpha$ is often critical to the performance of   learning algorithms.  In principle, we can tune algorithms for many $l_\alpha$ distances; and {\em stable random projections} can provide an efficient tool.

To recover  the original distances, we face an estimation task. Compared with previous estimators based on the {\em geometric mean}, {\em the harmonic mean}, or the {\em fractional power}, the proposed {\em optimal quantile} estimator exhibits two advantages. Firstly, the {\em optimal quantile} estimator is nearly one order of magnitude more efficient than other estimators (e.g., reducing the training time from one week to one day).  Secondly, the {\em optimal quantile} estimator is  considerably more accurate when $\alpha >1$, in terms of both the variances and error (tail) probabilities. Note that $\alpha \geq1$ corresponds to a convex norm (satisfying the triangle inequality), which might be another motivation for using $l_\alpha$ distances with $\alpha \geq1$.

One theoretical contribution is the explicit tail bounds for general quantile estimators and consequently the  sample complexity bound $k = O\left(\log n/\epsilon^2\right)$. Those bounds  may guide practitioners in choosing $k$, the number of projections. The (practically useful) bounds are expressed in terms of the probability functions and hence they might be not as convenient for further theoretical analysis. Also, we should mention that the bounds do not recover the optimal bound of the {\em arithmetic mean} estimator when $\alpha = 2$, because the {\em arithmetic mean} estimator is statistically optimal at $\alpha =2$ but the {\em optimal quantile} estimator is not.

While we believe that applying {\em stable random projections} in machine learning has become straightforward, there are interesting theoretical issues for future research. For example, how theoretical properties of learning algorithms may be affected if the approximated (instead of exact) $l_\alpha$ distances are used?

\appendix
\vspace{-0.1in}
\section{Proof of Lemma \ref{lem_var_q}}\label{proof_lem_var_q}
\vspace{-0.05in}

Denote $f_X\left(x; \alpha,d_{(\alpha)}\right)$ and $F_X\left(x; \alpha,d_{(\alpha)}\right)$ the probability density function and the cumulative density function  of $X\sim S(\alpha, d_{(\alpha)})$, respectively. Similarly we use $f_Z\left(z; \alpha,d_{(\alpha)}\right)$ and $F_Z\left(z; \alpha,d_{(\alpha)}\right)$ for $Z = |X|$. Due to  symmetry, the following relations hold
{\small\begin{align}\notag
&f_Z\left(z;\alpha,d_{(\alpha)}\right) = 2f_X\left(z;\alpha,d_{(\alpha)}\right) = 2/{d_{(\alpha)}^{1/\alpha}}f_X\left(z/{d_{(\alpha)}^{1/\alpha}};\alpha,1\right),\\\notag
&F_Z\left(z;\alpha,d_{(\alpha)}\right) = 2F_X\left(z;\alpha,d_{(\alpha)}\right)-1 = 2F_X\left({z}/{d_{(\alpha)}^{1/\alpha}};\alpha,1\right)-1,\\\notag
&F_Z^{-1}\left(q;\alpha,d_{(\alpha)}\right) = F_X^{-1}\left((q+1)/2;\alpha,d_{(\alpha)}\right) = d_{(\alpha)}^{1/\alpha}F_X^{-1}\left((q+1)/2;\alpha,1\right).
\end{align}}\vspace{-0.2in}

Let {\small$W =q\text{-Quantile}\{|S(\alpha,1)|\}= F_X^{-1}\left((q+1)/2;\alpha,1\right)$} and $W_d = F^{-1}_Z\left(q;\alpha,d_{(\alpha)}\right) = d_{(\alpha)}^{1/\alpha} W$. Then, following known statistical results, e.g., \cite[Theorem 9.2]{Book:David}, the asymptotic variance of $\hat{d}_{\alpha,q}^{1/\alpha}$ should be
{\small\begin{align}\notag
\text{Var}\left(\hat{d}_{\alpha,q}^{1/\alpha}\right)=& \frac{1}{k}\frac{q-q^2}{f^2_Z\left(W_d;\alpha,d_{(\alpha)}\right) W^2 } + O\left(\frac{1}{k^2}\right)
=\frac{1}{k}\frac{q-q^2}{d_{(\alpha)}^{-2/\alpha}f^2_Z\left(W;\alpha,1\right) W^2} + O\left(\frac{1}{k^2}\right)\\\notag
=&\frac{1}{k}\frac{q-q^2}{4d_{(\alpha)}^{-2/\alpha}f^2_X\left(W;\alpha,1\right) W^2} + O\left(\frac{1}{k^2}\right).
\end{align}}
By ``delta method," i.e., {\small$\text{Var}\left(h(x)\right) \approx \text{Var}\left(x\right)\left(h^\prime(x)\right)^2$},
{\small\begin{align}\notag
\text{Var}\left(\hat{d}_{\alpha,q}\right)&= \text{Var}\left(\hat{d}_{\alpha,q}\right) \left(\alpha d_{(\alpha)}^{(\alpha-1)/\alpha}\right)^2 +  O\left(\frac{1}{k^2}\right)
=\frac{1}{k}\frac{(q-q^2)\alpha^2/4}{f^2_X\left(W;\alpha,1\right) W^2 } d_{(\alpha)}^2+O\left(\frac{1}{k^2}\right).
\end{align}}

\vspace{-0.3in}
\section{Proof of Lemma \ref{lem_convexity}}\label{proof_lem_convexity}
\vspace{-0.05in}
First, consider $\alpha = 1$. In this case,
{\small\begin{align}\notag
&f_X(x;1,1) = \frac{1}{\pi}\frac{1}{x^2+1}, \hspace{0.3in} W = F_X^{-1}\left((q+1)/2;1,1\right) = \tan\left(\frac{\pi}{2}q\right),\\\notag
&g(q;1)  = \frac{q-q^2}{\left(\frac{2}{\pi}\frac{1}{\tan^2\left(\frac{\pi}{2}q\right)+1}\right)^2\tan^2\left(\frac{\pi}{2}q\right)}
=\frac{q-q^2}{\sin^2(\pi q)}\pi^2.
\end{align}}
It suffices to study {\small$L(q) = \log g(q;1)$.}
{\small\begin{align}\notag
&L^\prime(q) = \frac{1}{q} - \frac{1}{1-q} -\frac{2\pi\cos(\pi q)}{\sin(\pi q)},\hspace{0.3in} L^{\prime\prime}(q) = -\frac{1}{q^2} - \frac{1}{(1-q)^2} + \frac{2\pi^2}{\sin^2(\pi q)}.
\end{align}}
Because $\sin(x)\leq x$ for $x\geq 0$, it is easy to see that {\small$\frac{\pi}{\sin(\pi q)} - \frac{1}{q}\geq 0$, and  $\frac{\pi}{\sin(\pi q)} - \frac{1}{1-q} =\frac{\pi}{\sin(\pi (1-q))} - \frac{1}{1-q} \geq 0$}.
Thus, $L^{\prime\prime}\geq 0$, i.e., $L(q)$ is convex and so is $g(q;1) = e^{L(q)}$.  Since $L^\prime(1/2) = 0$, we know $q^*(1) = 0.5$.\\

Next we consider $\alpha = 0+$, using a fact \cite{Proc:Li_SODA08} that as $\alpha \rightarrow 0+$, $|S(\alpha,1)|^\alpha$ converges to $1/E_1$, where $E_1$ stands for an exponential distribution with mean 1.

Denote $h = d_{(0+)}$ and $z_j \sim h/E_1$. The sample quantile estimator becomes
{\small\begin{align}\notag
\hat{d}_{(0+),q} = \frac{q\text{-Quantile}\{|z_j|, j = 1, 2, ..., k\}}{q\text{-Quantile}\{1/E_1\}}.
\end{align}}
In this case,\vspace{-0.2in}
{\small\begin{align}\notag
&f_Z(z;h) = e^{-h/z}\frac{h}{z^2},  \hspace{0.1in} F_Z^{-1}(q;h) = -\frac{h}{\log q},\\\notag
&\text{Var}\left(\hat{d}_{(0+),q}\right)
=\frac{1}{k}\frac{1-q}{q\log^2q}h^2+O\left(\frac{1}{k^2}\right).
\end{align}}
It is straightforward to show that $\frac{1-q}{q\log^2q}$ is a convex function of $q$ and the minimum is attained by solving $-\log q^* + 2q^* - 2 = 0$, i.e., $q^* = 0.203$.

\vspace{-0.1in}
\section{Proof of Lemma \ref{lem_bounds}}\label{proof_lem_bounds}
\vspace{-0.05in}
Given $k$ i.i.d. samples, $x_j\sim S(\alpha,d_{(\alpha)})$, $j = 1$ to $k$. Let $z_j = |x_j|$, $j = 1$ to $k$.
Denote by $F_Z(t;\alpha,d_{(\alpha)})$ the cumulative
density of $z_j$, and by $F_{Z,k}(t;\alpha,d_{(\alpha)})$ the empirical cumulative
density of $z_j$, $j = 1$ to $k$.

It is the basic fact\cite{Book:David} about order statistics that  $kF_{Z,k}(t;\alpha,d_{(\alpha)})$ follows a binomial, i.e.,  $kF_{Z,k}(t;\alpha,d_{(\alpha)})\sim Bin(k, F_Z(t;\alpha,d_{(\alpha)}))$.   For simplicity, we replace $F_{Z}(t;\alpha, d_{(\alpha)})$ by $F(t,d)$, $F_{Z,k}(t;\alpha, d_{(\alpha)})$ by $F_k(t,d)$, and $d_{(\alpha)}$ by $d$,  in this proof.

Using the  {\em original} binomial Chernoff bounds
\cite{Article:Chernoff_52},  we obtain, for $\epsilon^\prime>0$,
{\scriptsize\begin{align}\notag
&\mathbf{Pr}\left(kF_k(t;d)\geq(1+\epsilon^\prime)
  kF(t;d)\right)\\\notag \leq&\left(\frac{k -
    kF(t;d)}{k-(1+\epsilon^\prime)kF(t;d)}\right)^{k - k(1+\epsilon^\prime)F(t;d)}
\left(\frac{kF(t;d)}{(1+\epsilon^\prime)kF(t;d)}\right)^{(1+\epsilon^\prime)kF(t;d)}\\\notag
=&\left[\left(\frac{1 -F(t;d)}{1-(1+\epsilon^\prime)F(t;d)}\right)^{1 - (1+\epsilon^\prime)F(t;d)}
\left(\frac{1}{1+\epsilon^\prime}\right)^{(1+\epsilon^\prime)F(t;d)}\right]^k,
\end{align}}
and for $0<\epsilon^\prime<1$,
{\scriptsize\begin{align}\notag
&\mathbf{Pr}\left(kF_k(t;d)\leq(1-\epsilon^\prime)kF(t;d)\right)
\\\notag \leq &
\left[\left(\frac{1 -F(t;d)}{1-(1-\epsilon^\prime)F(t;d)}\right)^{1 - (1-\epsilon^\prime)F(t;d)}
\left(\frac{1}{1-\epsilon^\prime}\right)^{(1-\epsilon^\prime)F(t;d)}\right]^k.
\end{align}}\vspace{-0.15in}

Consider the general quantile estimator $\hat{d}_{(\alpha),q}$ defined in (\ref{eqn_quantile}).
For $\epsilon>0$, (again, denote  $W = q\text{-quantile}\{|S(\alpha, 1)|\}$),
{\scriptsize\begin{align}\notag
&\mathbf{Pr}\left( \hat{d}_{(\alpha),q} \geq(1+\epsilon)
  d\right)
= \mathbf{Pr}\left(q\text{-quantile}\{|x_j|\}\right) \geq ((1+\epsilon)d)^{1/\alpha}W)\\\notag
   =&  \mathbf{Pr}\left(  k
  F_k\left(\left(1+\epsilon\right)^{1/\alpha}W;1\right)\leq qk  \right)
=
  \mathbf{Pr}\left(kF_k(t;1)\leq(1-\epsilon^\prime)kF(t;1)\right),
\end{align}}
where  { $t = \left(1+\epsilon\right)^{1/\alpha}W$ and $q= (1-\epsilon^\prime)F(t;1)$}. Thus
{\scriptsize\begin{align}\notag
&\mathbf{Pr}\left( \hat{d}_{(\alpha),q} \geq(1+\epsilon)  d\right) \\\notag
\leq&
\left[\left(\frac{1-F\left( \left((1+\epsilon)\right)^{1/\alpha}W;1\right)}{1-q}\right)^{1-q}\left(\frac{F\left(\left((1+\epsilon)\right)^{1/\alpha}W;1\right)}{q}\right)^{q}
\right]^k
=\exp\left(-k\frac{\epsilon^2}{G_{R,q}}\right).
\end{align}}
\noindent where \vspace{-0.1in}
{\small\begin{align}\notag
&\frac{\epsilon^2}{G_{R,q}} = -(1-q)\log\left(1-F\left( \left(1+\epsilon\right)^{1/\alpha}W;1\right)\right)\\\notag
& - q \log \left(F\left( \left(1+\epsilon\right)^{1/\alpha}W;1\right)\right) + (1-q)\log (1-q) + q\log(q).
\end{align}}
For $0<\epsilon<1$,
{\small\begin{align}\notag
&\mathbf{Pr}\left( \hat{d}_{(\alpha),q} \leq (1-\epsilon)
  d\right) = \mathbf{Pr}\left(  k
  F_k\left(\left(1-\epsilon\right)^{1/\alpha}W;1\right)\geq qk \right)
= \mathbf{Pr}\left(kF_k(t;1)\geq(1+\epsilon^\prime)kF(t;1)\right),
\end{align}}
where $t = \left(1-\epsilon\right)^{1/\alpha}W$ and $q = (1+\epsilon^\prime)F(t;1)$. Thus,
{\scriptsize\begin{align}\notag
&\mathbf{Pr}\left( \hat{d}_{(\alpha),q} \leq (1-\epsilon) d\right) \\\notag
\leq&
\left[\left(\frac{1-F\left( \left(1-\epsilon\right)^{1/\alpha}W;1\right)}{1-q}\right)^{1-q}\left(\frac{F\left(\left(1-\epsilon\right)^{1/\alpha}W;1\right)}{q}\right)^{q}
\right]^k
=\exp\left(-k\frac{\epsilon^2}{G_{L,q}}\right),
\end{align}} where\vspace{-0.1in}
{\small\begin{align}\notag
&\frac{\epsilon^2}{G_{L,q}} = -(1-q)\log\left(1-F\left( \left(1-\epsilon\right)^{1/\alpha}W;1\right)\right)\\\notag
 &- q \log \left(F\left( \left(1-\epsilon\right)^{1/\alpha}W;1\right)\right) + (1-q)\log (1-q) + q\log(q).
\end{align}}\vspace{-0.1in}

Denote $f(t;d) = F^\prime(t;d)$. Using L'Hospital's rule
{\scriptsize\begin{align}\notag
&\underset{\epsilon\rightarrow0+}{\lim} \frac{1}{G_{R,q}}
 =
\underset{\epsilon\rightarrow0+}{\lim} \frac{-(1-q)\log\left(1-F\left( \left(1+\epsilon\right)^{1/\alpha}W;1\right)\right)}
{\epsilon^2}\\\notag
&\hspace{0.2in} +\frac{ - q \log \left(F\left( \left(1+\epsilon\right)^{1/\alpha}W;1\right)\right) + (1-q)\log (1-q) + q\log(q)}
{\epsilon^2}
\\\notag
=&\underset{\epsilon\rightarrow0+}{\lim} \frac{ f\left( \left(1+\epsilon\right)^{1/\alpha}W;1\right) \frac{W}{\alpha}\left(1+\epsilon\right)^{1/\alpha-1}}{F\left( \left(1+\epsilon\right)^{1/\alpha}W;1\right)\left(1-F\left( \left(1+\epsilon\right)^{1/\alpha}W;1\right)\right) }\times
\frac{F\left( \left(1+\epsilon\right)^{1/\alpha}W;1\right)-q}{2\epsilon}\\\notag
=&\underset{\epsilon\rightarrow0+}{\lim} \frac{ \left(f\left( \left(1+\epsilon\right)^{1/\alpha}W;1\right) \frac{W}{\alpha}\left(1+\epsilon\right)^{1/\alpha-1}\right)^2   }{2 F\left( \left(1+\epsilon\right)^{1/\alpha}W;1\right)\left(1-F\left( \left(1+\epsilon\right)^{1/\alpha}W;1\right)\right)  }\\\notag
=&\frac{f^2\left(W;1\right)W^2}{2q(1-q)\alpha^2}, \hspace{0.5in} (q = F(W,1)).
\end{align}}
Similarly\vspace{-0.2in}
{\small\begin{align}\notag
&\underset{\epsilon\rightarrow0+}{\lim} G_{L,q} = \frac{2q(1-q)\alpha^2}{f^2\left(W;1\right)W^2}.
\end{align}}\vspace{-0.15in}

\noindent To complete the proof, apply the relations on {\small$Z=|X|$} in the proof of Lemma \ref{lem_var_q}.


\end{document}